\documentclass[lettersize,journal]{IEEEtran}
\usepackage{amsmath}
\usepackage{amssymb}
\usepackage{amsfonts}
\usepackage[ruled,linesnumbered]{algorithm2e}
\usepackage{array}
\usepackage{subfigure}
\usepackage{textcomp}
\usepackage{stfloats}
\usepackage{url}
\usepackage{verbatim}
\usepackage{graphicx}
\usepackage{cite}
\usepackage{mathtools}
\usepackage{booktabs}
\usepackage{soul}
\SetKwInOut{Input}{Input}\SetKwInOut{Output}{Output}\SetKwInOut{Parameter}{Parameter}

\begin{document}

\title{Noise-Resistant Deep Metric Learning with Probabilistic Instance Filtering}

\author{Chang~Liu,
        Han~Yu$^*$,
        Boyang~Li$^*$,
        Zhiqi~Shen$^*$,
        Zhanning~Gao,
        Peiran~Ren,
        Xuansong~Xie,
        Lizhen~Cui
\thanks{C. Liu, H. Yu, B. Li and Z. Shen are with the School of Computer Science and Engineering, Nanyang Technological University, Singapore.}
\thanks{Z. Gao, P. Ren and X. Xie are with the Alibaba Group, Hangzhou, China.}
\thanks{L. Cui is with the School of Software, Shandong University, Jinan, China.}
\thanks{*Corresponding Authors: \{han.yu, zqshen, boyang.li\}@ntu.edu.sg}
}

\markboth{Journal of \LaTeX\ Class Files,~Vol.~14, No.~8, August~2021}%
{Shell \MakeLowercase{\textit{et al.}}: A Sample Article Using IEEEtran.cls for IEEE Journals}


\maketitle

\begin{abstract}
Noisy labels are commonly found in real-world data, which cause performance degradation of deep neural networks. Cleaning data manually is labour-intensive and time-consuming. Previous research mostly focuses on enhancing classification models against noisy labels, while the robustness of deep metric learning (DML) against noisy labels remains less well-explored. In this paper, we bridge this important gap by proposing Probabilistic Ranking-based Instance Selection with Memory (PRISM) approach for DML. PRISM calculates the probability of a label being clean, and filters out potentially noisy samples. 
Specifically, we propose a novel method, namely the von Mises-Fisher Distribution Similarity (vMF-Sim), to calculate this probability by estimating a von Mises-Fisher (vMF) distribution for each data class. Compared with the existing average similarity method (AvgSim), vMF-Sim considers the variance of each class in addition to the average similarity. With such a design, the proposed approach can deal with challenging DML situations in which the majority of the samples are noisy. 
Extensive experiments on both synthetic and real-world noisy dataset show that the proposed approach achieves up to 8.37\% higher Precision@1 compared with the best performing state-of-the-art baseline approaches, within reasonable training time.
\end{abstract}


\section{Introduction}
\IEEEPARstart{A}{s} a result of human annotation errors or imperfect automated data collection procedures, noisy labels are commonly found in real-world data. Noisy training data degrades the predictive power of neural networks \cite{han2018co,wang2018iterative,jiang2020beyond}. Manual inspection and correction of labels are labour-intensive, and hence scale poorly to large datasets. Consequently, real-world applications of deep neural network call for robust training techniques that can maintain model performance in the presence of label noise.
Existing research effort on noise-resistant learning has mostly been focusing on classification tasks \cite{patrini2017making,han2018co,yu2019does,li2019learning,li2020dividemix,jiang2018mentornet,wang2018iterative,wei2020combating,jiang2020beyond}. Noise-resistant deep metric learning (DML) remains an open research area, despite its vulnerability to label noise \cite{wang2017robust}. In this paper, we set out to bridge this important gap in existing literature.

The aim of DML is to learn a distance metric between data samples, which can be in the form of images, videos or text snippets. After training, the learned metric shall return small distances among data from the same class and large distances among data from different classes. DML plays important roles in a diverse set of applications including image/video retrieval \cite{kaya2019deep,revaud2019learning,gordo2016deep,avgoustinakis2020audio,shao2021temporal}, person re-identification \cite{wang2018learning,luo2019bag,ye2021deep,zhang2020ordered,chen2017person}, vehicle re-identification \cite{Wang_2017_ICCV,meng2020parsing}, and self-supervised learning \cite{miech2020end,he2020momentum,chen2020simple,grill2020bootstrap}. 
It differs from traditional supervised classification in three main aspects. Firstly, the definition of classes in DML can be even more fine-grained than typical fine-grained classification tasks. For example, in the Stanford Online Products dataset \cite{oh2016deep}, every furniture item on Ebay forms its own class. Secondly, DML does not assume a fixed set of classes. This reflects realistic scenarios in which new classes, such as new furniture items and new food recipes \cite{lee2017cleannet}, routinely emerge. Thirdly, for the above reasons, the learned metric is often tested on classes that do not exist in the training set. This contrasts with supervised classification, where the same classes are used for both training and testing. As a result, DML is a distinct problem from classification with a diverse set of techniques being proposed so far \cite{hermans2017defense,schroff2015facenet,yuan2017hard,cakir2019deep,wang2020cross,brown2020smooth}.


Following \cite{wang2019multi,wang2020cross}, DML can be categorized into pair-based methods \cite{chopra2005learning,hadsell2006dimensionality,hoffer2015deep,schroff2015facenet,wang2019ranked,wang2019multi,sun2020circle} and proxy-based methods \cite{cakir2019deep,zhaiclassification,kim2020proxy,movshovitz2017no,qian2019softtriple}. Neither category is immune to label noise. Pair-based methods compute the loss with pair-wise similarities between individual data points, for which label noise may cause incorrect data pairing. Proxy-based methods compare the data points against proxies to calculate the loss function. A proxy is a trainable vector that represents a group of data points. With label noise, the learned proxy may deviate from the true center.


Currently, the only DML method capable of handling label noise is the Probabilistic Ranking-based Instance Selection with Memory (PRISM) approach proposed in our previous work \cite{liu2021noise}. PRISM calculates the probability that a data point $i$ has a correct label ($P_{\text{clean}}(i)$), and considers it noisy if $P_{\text{clean}}(i)$ falls below a threshold $m$. It leverages Average Similarity (AvgSim), which computes $P_{\text{clean}}(i)$ based on the similarity between potentially noisy data and clean data. The major drawback in this approach is that the threshold $m$ is set in an arbitrary manner without regard to the context of the underlying class.

In this paper, we extend PRISM by introducing a new technique for computing $P_{\text{clean}}(i)$, namely the von Mises-Fisher distribution Similarity (vMF-Sim). It estimates a von Mises-Fisher distribution \cite{mardia2009directional} for each cluster and computes $P_{\text{clean}}$ in a Bayesian manner. Compared with avgSim, vMF-Sim considers the variance within each class in addition to the average similarity instead of relying on an arbitrary threshold, making it more aware of the context of each class. In addition, a memory bank is incorporated to store previously computed data representations to increase the number of samples available for noise filtering. This technique reduces the variance of $P_{\text{clean}}(i)$ estimation.

To evaluate the proposed techniques, we conduct extensive experiments on both synthetic and real-world noisy datasets. We synthesize noisy labels on commonly used DML datasets (e.g., CARS \cite{krause20133d}, CUB \cite{wah2011caltech} and SOP \cite{oh2016deep}) under label noise rate settings ranging from 10\% to 75\%. We also conduct experiments on real-world noisy datasets, FOOD-101N \cite{lee2017cleannet} and CARS-98N. We use a separate validation set to perform model selection. The results show that PRISM significantly outperforms the baseline methods by up to 8.37\% in terms of Precision@1 score. The Precision@1 score is boosted by 5.16\% under PRISM+vMF-Sim compared to the original approach using AvgSim. The proposed approaches offer a useful tool to enable DML to operate in the presence of label noise, making DML technique more robust for practical applications.

\section{Related Works}

\subsection{Noise-resistant Training in Classification} Training under noisy labels has been studied extensively for classification tasks \cite{angluin1988learning,wang2018iterative,wei2020combating,jiang2020beyond,zheng2021mlc,algan2020meta}. A common approach is to detect noisy labels and exclude them from the training set. Recent works have also started exploring correcting the noisy labels \cite{zheng2021mlc,algan2020meta} or treating noisy data as unlabeled data for semi-supervised learning \cite{li2020dividemix}.

F-correction \cite{patrini2017making} models label noise with a class transition matrix. MentorNet \cite{jiang2018mentornet} trains a teacher network that provides a weight for each sample to the student network. Co-teaching \cite{han2018co} trains two networks concurrently. Samples identified as having small loss by one network is used to train the other network. This is further improved in \cite{yu2019does} by training on samples that have small losses and different predictions from the two networks. Many recent works \cite{yu2019does,li2020dividemix,liu2020early,nishi2021improving} 
filter noisy data based on the predicted class output by the current classification model. Nevertheless, few existing work focuses on addressing the label noise problem in DML. 

\subsection{Deep Metric Learning} 
We broadly categorize deep metric learning into: 1) pair-based methods, and 2) proxy-based methods. 
Pair-based methods \cite{schroff2015facenet,wang2019ranked,wang2019multi,sun2020circle} calculate the loss based on the contrast between positive pairs and negative pairs. Commonly used loss functions include contrastive loss \cite{chopra2005learning}, triplet loss \cite{hermans2017defense} and softmax loss \cite{goldberger2004neighbourhood}. In this process, identifying informative positive and negative pairs becomes an important consideration \cite{Song2016:lifted-embedding,harwood2017smart,Suh_2019_CVPR,cakir2019deep,wang2019multi,sun2020circle,wu2017sampling}. The problem of overfitting in DML has also been considered \cite{opitz2018deep}.  Proxy-based methods \cite{movshovitz2017no,qian2019softtriple,cakir2019deep,kim2020proxy,teh2020proxynca++,zhu2020fewer,elezi2020group,zhaiclassification,boudiaf2020unifying} represent each class 
with one or more proxy vectors, and use the similarities between the input data and the proxies to calculate the loss. Proxies are learned from data during model training, which could deviate from the class center under heavy noise and cause performance degradation. 


\subsection{Noisy Labels in Metric Learning} 
To our knowledge, the only method which explicitly handles noise in neural metric learning is \cite{wang2017robust}. The technique estimates the posterior label distribution using a shallow Bayesian neural network with only one layer. Due to its computational complexity, the approach may not scale well to deeper network architectures.

A few works attempt to handle outliers in DML training data, but do not explicitly deal with substantial label noise. In \cite{wang2019deep}, the pair-based loss is used to train a proxy for each class simultaneously to adjust the weights of the outliers, but substantial label noise may cause the learned proxies to be inaccurate. In \cite{ozaki2019large}, noisy data in DML in handled by performing label cleaning using a model trained on a clean dataset. However, such a dataset may not be available in real-world applications. In this paper, we do not rely on the existence of a clean dataset.


\section{The PRISM Approach}
The detection of noisy data labels usually attempts to identify data points that stand out from others in the same class. On the one hand, distinguishing noisy data samples from the rest requires an effective similarity metric. On the other hand, learning a good similarity metric depends on the availability of clean training data. 

To resolve this dilemma, PRISM adopts an online approach that performs metric learning and noisy data filtering simultaneously. The general idea is, for every data point, PRISM estimates the probability that it is clean using the currently learned similarity metric. If the probability exceeds a threshold, the data point is used to update the network and the similarity metric. Otherwise, it is discarded. 

\subsection{Algorithm Outline}
We start with introducing the notations. The training set of $N$ data points is denoted as $\mathcal{X}=\{(x_0,y_0), (x_1,y_1), ...,$ $(x_N,y_N)\}$, where $x_i$ is an input data and $y_i$ is the given class label. In deep metric learning, we learn a function, $f(\cdot)$, that outputs a feature vector $f(x_i)$ for $x_i$. An ideal $f(\cdot)$ should produce high similarity between the two feature vectors $f(x_i)$ and $f(x_j)$ if they have the same label (i.e. $y_i=y_j$); otherwise, the similarity should be low. 
In this paper, we adopt the cosine similarity, defined as
\begin{equation}\label{equ:sim}
S(f(x_i), f(x_j))=\frac{f(x_i)^Tf(x_j)}{\|f(x_i)\| \|f(x_j)\|}.
\end{equation}
We denote the current mini-batch of $B$ data points as $\mathcal{B}=\{(x_0,y_0),...,(x_B,y_B)\}$. A pair of data points $(x_i, x_j)$ is a positive pair if they have the same label and a negative pair otherwise.

\begin{algorithm}[t!]
	\SetAlgoLined
	\Input{Training set $\mathcal{X}$; Batch size $B$;\\Loss function $L(\cdot)$;\\Total number of training iteration $I_{total}$;\\Initial neural network $f(\cdot)$;}
	\BlankLine
	\If {using AvgSim or vMF-Sim}{
	Memory $\mathcal{M} \gets$  an empty queue;\\}
	\While {$I_{total}$ not reached}{
	Sample a minibatch $\mathcal{B}=\{(x_0,y_0),...,(x_B,y_B)\}$\\
	Calculate $l_2$ normalized $f(\mathcal{B})$\\ 
	\For{\textup{\textbf{each}} $(x_i,y_i) \in \mathcal{B}$}{
	    \eIf{we observe class $y_i$ for the first time}{
	    $P_{\text{clean}}(i) \gets 1$;\\
	    }
	    {
	    Calculate $P_{\text{clean}}(i)$ using AvgSim, ProxySim, or vMF-Sim (Sec~\ref{sec:avgsim}-\ref{sec:vmfsim});
	    }
	}
	Calculate the threshold $m$ (Sec~\ref{sec:threshold});\\
	$\mathcal{B}_{clean} \gets \varnothing$\\
	\For{\textup{\textbf{each}} $(x_i,y_i) \in \mathcal{B}$}{
	    \If{$P_{\text{clean}}(i)>m$}
	    {
	        Add $(f(x_i),y_i)$ to $\mathcal{B}_{clean}$;\\
	    }
	}
	\If {using AvgSim or vMF-Sim}{
	Enqueue $\mathcal{B}_{clean}$ into $\mathcal{M}$;\\}
	Calculate loss $L(\mathcal{B}_{clean})$ and update the parameters of $f(\cdot)$;\\
	}
	\BlankLine
	\caption{The PRISM algorithm}\label{alg:nsm}
\end{algorithm}

PRISM adopts an online data filtering approach, shown as Algorithm~\ref{alg:nsm}. At every training iteration, we sample $B$ data as a minibatch, where $B$ is the batch size (Line 5). Specifically, we sample $P$ unique classes and $K$ images for each selected class, so that $B=PK$. We then calculate the probability of $y_i$ being correct $P_{\text{clean}}(i)$ for each $(x_i,y_i)$ tuple using any of the techniques discussed in Sections \ref{sec:avgsim}-\ref{sec:vmfsim}. 

If a class $y_i$ is encountered for the first time, all data samples in class $y_i$ are considered clean (i.e. set $P_{\text{clean}}(i)=1$, Line 9), since we have no knowledge on the data distribution of class $y_i$. If the class has been observed before, we will estimate $P_{\text{clean}}(i)$ (Line 11 of Algorithm~\ref{alg:nsm}). 

After that, we compute the threshold $m$ using techniques described in Section~\ref{sec:threshold}. If $P_{\text{clean}}(i)$ falls under $m$, we consider the corresponding data point to be clean and calculate loss on the clean data only. In AvgSim and vMF-Sim, we store all clean ($l_2$ normalized) features in a memory bank (Line 18), which is later used in the calculation of $P_{\text{clean}}$. 
Finally, the loss (Section~\ref{sec:loss}) is calculated to update parameters of the model $f(\cdot)$ (Line 24 of Algorithm~\ref{alg:nsm}). We compute the loss using only clean data and perform stochastic gradient descent.

The PRISM algorithm relies on the estimate of $P_{\text{clean}}(i)$, or the probability that a data point $(x_i, y_i)$ is clean. In the following sections, we describe three methods to estimate $P_{\text{clean}}(i)$, namely Average Similarity (AvgSim), Proxy Similarity (ProxySim) and von Mises-Fisher Distribution Similarity (vMF-Sim).

\subsection{Average Similarity}
Average Similarity (AvgSim) compares the features of $x_i$ with the content of the memory bank to determine if the label $y_i$ is noisy. 

\textbf{The memory bank} is a first-in-first-out queue $\mathcal{M}=\{(v_0,y_0),(v_1,y_1),...,(v_M,y_M)\}$, that stores features of historical data samples. $M$ is the size of the memory bank. In the beginning, the memory bank is initialized as an empty queue. After a clean minibatch data $\mathcal{B}_{clean}$ are identified, we append the current feature $v_i=f(x_i)$ of the clean data $x_i$ to the memory bank for calculating $P_{\text{clean}}(i)$ in future iterations. If the maximum bank capacity is reached, the oldest features are dequeued from the memory bank, so that the more recent features are kept.

If $y_i$ is a clean label, then the similarity between $x_i$ and same-class data representations in the memory should be larger compared to its similarity with representations from other classes. Based on this intuition, we define
\begin{equation}\label{equ:pi}
P_{\text{clean}}(i) = \frac{\exp\left(T(x_i, y_i)\right)}{\sum_{k \in C}\exp \left(T(x_i, k\right))}
\end{equation}
\begin{equation}
T(x_i, k) = \frac{1}{M_k}\sum\limits_{(v_j,y_j) \in \mathcal{M}, y_j=k}S(f(x_i),v_j)
\end{equation}
Here, $M_k$ is the number of class $k$ samples in the memory bank. $T(x_i, k)$ is the average similarity between $x_i$ and all the stored features $v_j$ of class $k$. From Bayesian statistics, we know that
\begin{equation}\label{equ:bayesian}
P(y=k|X=x_i) = \frac{P(X=x_i|Y=k)P(Y=k)}{\sum_{k' \in C} P(X=x_i|Y=k')P(Y=k')}
\end{equation}
Comparing Eq.~\eqref{equ:pi} and \eqref{equ:bayesian}, we see that $\exp(T(x_i, k))$ can be treated as the probability $P(X=x_i|Y=k)$ up to a constant $z_k$, assuming a uniform prior $P(Y=k)$ and identical $z_k$ for every class. Although similar forms can be found in applications such as metric learning \cite{goldberger2004neighbourhood,qian2019softtriple}, data visualization \cite{vandermaaten2008:tsne} and uncertainty estimation \cite{mandelbaum2017distance,xing2020distance}, we note that its use in noise-resistant DML is novel. 

\subsubsection{Accelerating Average Similarity Calculation}\label{sec:avgsim}
The average similarity is computed from the similarities between all pairs of data samples, which can incur high computational overhead. Here, we propose a simple technique for improving its efficiency. For the $k^{\text{th}}$ cluster, we replace the $M_k$ data samples with the mean feature vector $w_k$ of the class,
\begin{align}
\begin{split}\label{equ:wk}
    \sum_{\substack{(v_j,y_j) \in \mathcal{M} \\ y_j=k}}\frac{S(f(x_i),v_j)}{M_k}&=(\frac{1}{M_k}\sum_{\substack{(v_j,y_j) \in \mathcal{M} \\ y_j=k}}\frac{v_j}{\|v_j\|})\frac{f(x_i)}{\|f(x_i)\|}\\ &=w_k\frac{f(x_i)}{\|f(x_i)\|},
\end{split}\\
\begin{split}\label{equ:wk_used_in_alg1}
    w_k &= \frac{1}{M_k} \sum_{(v_j,y_j) \in \mathcal{M},  y_j=k} \frac{v_j}{\|v_j\|}.
\end{split}
\end{align}
Substituting Eq.~\eqref{equ:wk} into Eq.~\eqref{equ:pi}, $P_{\text{clean}}(i)$ can be expressed as:
\begin{equation}\label{equ:pi2}
P_{\text{clean}}(i)=\exp\left(w_{y_i}\frac{f(x_i)}{\|f(x_i)\|}\right) / {\sum_{k \in C}\exp\left(w_k\frac{f(x_i)}{\|f(x_i)\|}\right)}.
\end{equation}
Eq.~\eqref{equ:pi2} requires a mean feature vector $w_k$ to be maintained for each class $k$. In the implementation, after we enqueue a batch of clean data $\mathcal{B}_{clean}$ into the memory bank, $w_k$ is updated for the class that appeared in $\mathcal{B}_{clean}$ using Eq.~\eqref{equ:wk_used_in_alg1}.

The time complexity of Eq.~\eqref{equ:pi} is $O(BM)$ for a minibatch of size $B$ and a memory bank with $M$ samples. By following Eq.~\eqref{equ:pi2}, the time complexity is reduced to $O(B|C|)$. $|C|$ is the total number of classes, which is much smaller than $M$. This technique accelerates the calculation of the average similarity by a factor of $\frac{M}{|C|}$.

\subsection{Proxy Similarity}
By extending Eq.~\eqref{equ:pi2}, we can replace $w_k$ with a proxy trained by a proxy-based method. This leads to another method to calculate $P_{\text{clean}}(i)$, referred to as the Proxy Similarity method (ProxySim).
\begin{equation}
P_{\text{clean}}(i)=\frac{\exp\left(S(p_{y_i}(x_i),f(x_i)\right)}{\sum_{k \in C}\exp\left(S(p_k(x_i), f(x_i)\right)},
\end{equation}
where $p_k(x_i)$ is the proxy of the class $k$ trained by a proxy-based method. We assume the proxy-based method maintains $H$ proxy vectors for each class, and selects the proxy having the maximum similarity with $f(x_i)$:
\begin{equation}
p_{k}(x_i)=\operatorname*{argmax}_{p_k^h}S(p_k^h,f(x_i)),
\end{equation}
where $h \in [1,H]$.

With ProxySim, the memory bank is no longer required for calculating $P_{\text{clean}}(i)$, thereby significantly reducing GPU memory usage. It is useful in application which require local memory consumption. 
Nevertheless, due to the need for training the proxy, ProxySim is only suitable for proxy-based methods. 

\subsection{von Mises-Fisher Distribution Similarity}\label{sec:vmfsim}
In AvgSim and ProxySim, different classes are assumed to have equal variance. Here, we introduce von Mises-Fisher Distribution Similarity (vMF-Sim), which relaxes this assumption and considers both the similarity and the variance within each class. 

The von Mises-Fisher distribution is a probability distribution defined on the $(D-1)$-hypersphere $\mathbb{S}^D$ \cite{mardia2009directional}. 
The probability density function (PDF) of the von Mises-Fisher distribution is defined as:
\begin{equation}\label{equ:vMF}
p^{\text{vMF}}(i)=C_D(\kappa)\exp(\kappa \mu^T x_i),
\end{equation}
where $\kappa \in \mathbb{R}_+$ is a non-negative parameter that describes the ``tightness'' of the distribution. Larger $\kappa$ means the distribution is more concentrated around the center $\mu \in \mathbb{S}^D$. When $\kappa=0$, the data are distributed uniformly on the sphere. The normalization term $C_D(\kappa)$ is computed as 
\begin{equation}\label{equ:C_D}
C_D(\kappa)=\frac{\kappa^{\frac{D}{2}-1}}{(2\pi)^{\frac{D}{2}}I_{\frac{D}{2}-1}(\kappa)},
\end{equation}
where $I_{\frac{D}{2}-1}(\cdot)$ is the modified Bessel function of the first kind at order $(\frac{D}{2}-1)$. 

The rationale behind adopting vMF is as follows. Gaussian distribution is commonly used in machine learning to model the feature distributions \cite{li2020dividemix,liu2021handwritten,viroli2019deep}. A feature (as a vector) can be divided into its direction (i.e., the vector on a unit sphere $\mathbb{S}^D$) and $l_2$ norm. The commonly-used cosine similarity (Eq.~\eqref{equ:sim}) requires an $l_2$ normalization on the feature, such that only the directions are considered during similarity calculation. The Gaussian distribution on $\mathbb{R}^D$ uses both the direction and norm, which does not align well with the cosine similarity. Thus, we use the Gaussian distribution on a sphere, which is known as the von Mises-Fisher (vMF) distribution \cite{mardia2009directional}, to model the $l_2$ normalized features. Another reason is that the features are always $l_2$ normalized before being added into the memory bank, such that the cosine similarity can be calculated by a dot product operation. Applying vMF on the features in the memory bank is more intuitive and easier than using the Gaussian distribution on $\mathbb{R}^D$.

We can estimate $\mu$ and $\kappa$ using maximum-likelihood estimation \cite{sra2012short,banerjee2005clustering} as
\begin{equation} \label{equ:vMF_mu}
\mu_k=\frac{\sum_{(v_i,y_i) \in \mathcal{M}, y_i=k}v_i}{\bar R},
\end{equation}
\begin{equation}\label{equ:vMF_kappa}
\kappa_k=\bar R \frac{D-\bar R}{1-\bar R^2},
\end{equation}
where $\bar R=\|\sum_{(v_i,y_i) \in \mathcal{M}, y_i=k}v_i\|$.

The estimated distribution $p^{\text{vMF}}_{k}(i)$ is the class-conditional probability $P(X=x_i|Y=k)P(Y=k)$. Assuming a uniform prior $P(Y=k)$, we have 
\begin{equation}\label{equ:vMF_pClean}
P_{\text{clean}}(i)=\frac{p(Y=y_i, X=x_i)}{\sum_{k \in C} p(Y=k, X=x_i)} = \frac{p^{\text{vMF}}_{y_i}(i)}{\sum_{k \in C}p^{\text{vMF}}_k(i)}.
\end{equation}

Similar to AvgSim, vMF-Sim updates the parameters $\mu_k$ and $\kappa_k$ using the clean data in the memory bank after every mini-batch. This requires an adequate number of clean samples in the memory bank. Therefore, in the first $I_{v}$ iterations, PRISM uses AvgSim to identify the clean samples. Afterwards, vMF-Sim replaces AvgSim. The hyperparameter $I_{v}$ can be determined through the model selection process.

\subsection{Setting the Threshold $m$}\label{sec:threshold}
When $P_{\text{clean}}(i)$ falls below a threshold $m$, we treat $(x_i, y_i)$ as a noisy data sample. 
We propose three methods to determine the value of the threshold $m$: 1) the fixed value method, 2) the top-$R$ method (TRM) and 3) the smooth top-$R$ method (sTRM). 

Setting $m$ as a fixed value for all iterations is easy to implement. However, it requires a clear gap between the $P_{\text{clean}}(i)$ value of noisy data and clean data. In the experiments, we show that only vMF-Sim is capable of providing such a clear and stable gap for this method to be useful.

Under TRM, we define a filtering rate (i.e. estimated noise rate) $R$. In each minibatch, we treat $(x_i, y_i)$ as noisy if $P_{\text{clean}}(i)$ falls within the lowest $R\%$ of all samples in the current minibatch $\mathcal{B}$. 

In contrast, sTRM keeps track of the average of the $R^{\text{th}}$ percentile of $P_{\text{clean}}(i)$ values over the last $\tau$ batches. Formally, let $Q_j$ be the $R^{\text{th}}$ percentile $P_{\text{clean}}(i)$ value in $j$-th mini-batch, the threshold $m$ is defined as:
\begin{equation}\label{equ:sTRM}
m=\frac{1}{\tau}\sum_{j=t-\tau}^{t}Q_j
\end{equation}
Compared to TRM, the sliding window approach of sTRM reduces the influence of a single mini-batch and creates a smoother and more accurate estimate of the $R^{\text{th}}$ percentile.





\subsection{Loss Functions}\label{sec:loss}
The traditional pair-based contrastive loss function \cite{chopra2005learning} computes similarities between all pairs of data samples within the mini-batch $\mathcal{B}$. The loss function encourages $f(\cdot)$ to assign small distances between samples in the same class and large distances between samples from different classes. 
More formally, the loss for mini-batch $\mathcal{B}$ is
\begin{equation}
\begin{split}
    L_{\text{batch}}(\mathcal{B}) =\sum_{\mathclap{\substack{(x_i, y_i), (x_j, y_j) \in B\\ y_i \neq y_j}}} [S(f(x_i), f(x_j))    -\lambda]_{+} -\sum_{\mathclap{\substack{(x_i, y_i), (x_j, y_j) \in B\\ y_i = y_j}}} S(f(x_i), f(x_j))
\end{split}
\end{equation}
where $\lambda\in[0,1]$ is a hyperparameter for the margin and $[x]_{+} = \max(x, 0)$.

With a memory bank $\mathcal{M}$ that stores the features of data samples in previous minibatches in a first-in-first-out manner \cite{wang2020cross}, we can employ many more positive and negative pairs in the loss, which may reduce the variance in the gradient estimates. The memory bank loss can be written as:
\begin{equation}
\begin{split}
    L_{\text{bank}}(\mathcal{M}, \mathcal{B}) =\sum_{\mathclap{\substack{(x_i,y_i)\in \mathcal{B},\\ (v_j,y_j) \in \mathcal{M},\\ y_i \neq y_j}}} \left[S(f(x_i), v_j) -\lambda\right]_{+} 
    -\sum_{\mathclap{\substack{(x_i,y_i)  \in \mathcal{B}, \\  (v_j,y_j) \in \mathcal{M}, \\ y_i = y_j}}} S(f(x_i),v_j).
    \end{split}
\end{equation}
The total loss is the sum of the batch loss $L_{\text{batch}}(\mathcal{B})$ and the memory bank loss $L_{\text{bank}}(\mathcal{M}, \mathcal{B})$, referred to as the \emph{memory-based contrastive loss} \cite{wang2020cross}. As we adopt the memory bank setup to identify data with noisy labels, PRISM works well under the memory-based contrastive loss. 

Another loss we employ with PRISM is the SoftTriple loss \cite{qian2019softtriple}, a type of proxy-based loss function. This loss maintains $H$ learnable proxies per class. A proxy is a vector that has the same size as the feature of an image. The similarity between an image to a given class of images is represented as a weighted similarity to each proxy in the class. The loss is computed as the similarities between the minibatch data and all classes:
\begin{equation}\label{eq:softtriple}
    L_{\text{SoftTriple}}=-\log\frac{\exp(\lambda(S'_{i,y_i}-\delta))}{\exp(\lambda(S'_{i,y_i}-\delta))+\exp(\lambda S'_{i,j})},
\end{equation}
\begin{equation}
    S'_{i,j}=\frac{\sum_{h=1}^H\exp\left(\gamma f(x_i)^\top p_j^h\right)f(x_i)^\top p_j^h}{\sum_{h=1}^H \exp\left(\gamma f(x_i)^\top p_j^h\right)}.
\end{equation}
$\lambda$ and $\gamma$ are predefined scaling factors. $\delta$ is a predefined margin. $p_j^h$ is the $h$-th proxy for class $j$, which is a learnable vector updated during model training.

\section{Experimental Evaluation}
To evaluate the effectiveness of the proposed PRISM approach on enabling DML to deal with noisy data, we compare it against 12 baseline approaches on both synthetic and real-world data. Further, we empirically analyze the reasons behind the performance of different variants of PRISM, and examine the contributions of different components of PRISM by ablation.
\subsection{Datasets}
\begin{figure}[t]
      \centering

      \includegraphics[width=1\columnwidth]{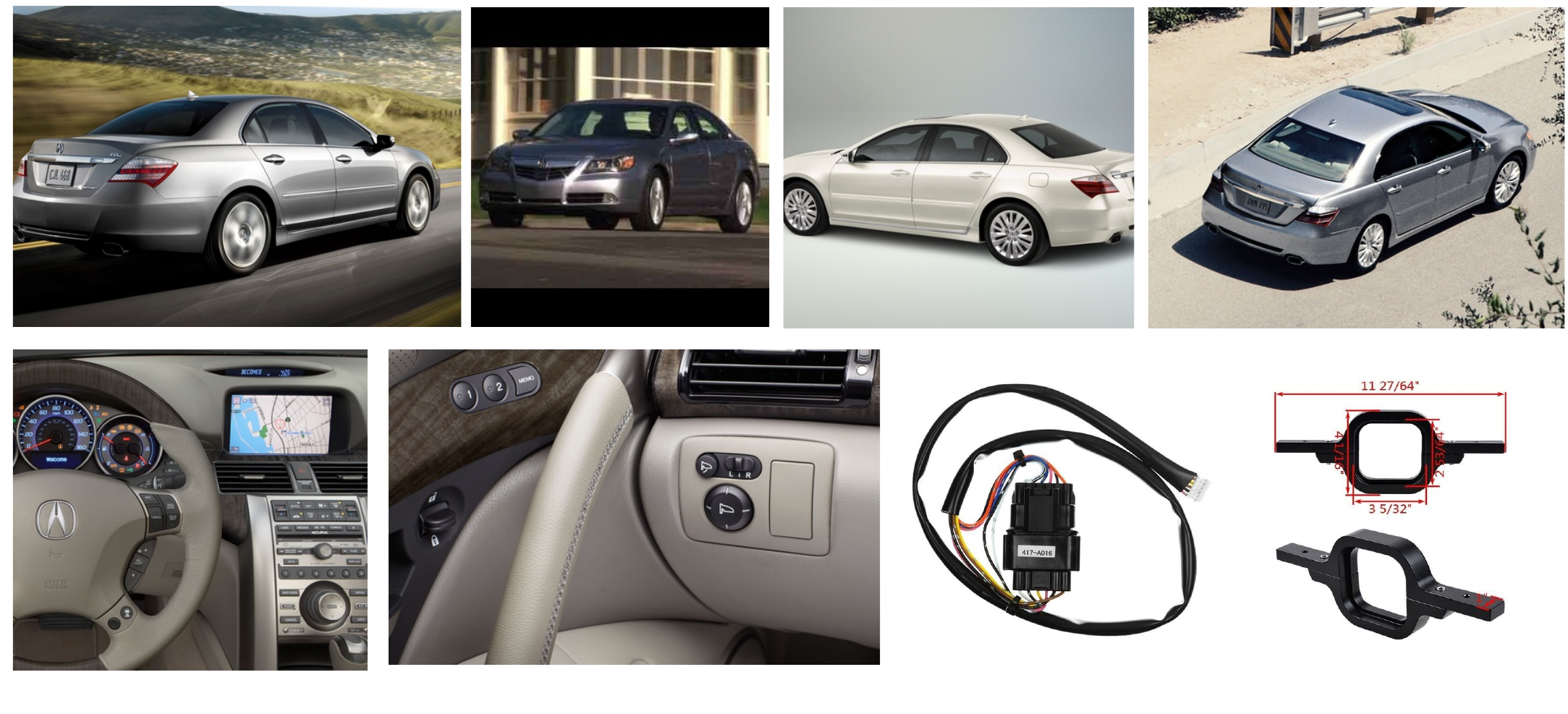}
      	\caption{Example images in CARS-98N. Images in the first row have clean labels. The second row shows some images with noisy labels, including car interiors and car parts.}\label{fig:carsn_example}
\end{figure}

We compare the algorithms on five datasets, including:
\begin{itemize}
    \item \textbf{CARS} \cite{krause20133d}, which contains 16,185 images of 196 different car models. We use the first 98 models for training, the next 48 models for evaluation and the last 48 models for testing, and incorporate synthetic label noise into the training set.
    \item \textbf{CUB} \cite{wah2011caltech}, which contains 11,788 images of 200 different bird species. We use the first 100 species for training, next 50 species for evaluation and the rest for testing, and incorporate synthetic label noise into the training set.
    \item \textbf{Stanford Online Products (SOP)} \cite{oh2016deep}, which contains 59,551 images of 11,318 furniture items on eBay. We use the first 50\% classes for training, next 25\% classes for validation and the rest for testing, and incorporate synthetic label noise into the training set.
    \item \textbf{Food-101N} \cite{lee2017cleannet}, a real-world noise dataset that contains 310,009 images of food recipes in 101 classes. It has the same 101 classes as Food-101 \cite{bossard14} (which is considered a clean dataset). We use 144,086 images belonging to the first 50 classes (in alphabetical order) as the training set, the next 25 classes with 25,000 images as the validation set, and the last 26 classes with 26,000 images as the test set.
    \item \textbf{CARS-98N} \cite{liu2021noise}, a real-world noise dataset that contains 9,558 images for 98 car models. The CARS-98N is \emph{only used for training}. Figure~\ref{fig:carsn_example} shows example images in this dataset. The noisy images often contain the interior of the car, car parts, or images of other car models. The validation and test set of CARS is used for model selection and performance evaluation.
\end{itemize}
Compared to the setting described in previous papers \cite{wang2019multi,wang2020cross,liu2021noise,wah2011caltech}, we use a validation set for model selection to avoid overfitting to the test set.

We adopt two models for noisy label synthesis: 1) symmetric noise and 2) Small Cluster noise. Symmetric noise \cite{van2015learning} has been widely used to evaluate the robustness of classification models \cite{han2018co,yu2019does,patrini2017making,li2019learning}. Given a clean dataset, the symmetric noise model assigns a predefined portion of data from every ground-truth class to all other classes with equal probability, without regard to the similarity between data samples. After the noise synthesis, the number of classes remains unchanged. 

Small Cluster noise mimics naturally occurring label noise. We observe that in real datasets, noisy data points tend to form small clusters. This is evident in Figure~\ref{fig:carsn_example}. The Small Cluster model flips the labels iteratively. In each iteration, we first cluster images from a randomly selected ground-truth class into a large number of small clusters. Each cluster is then merged into a randomly selected ground-truth class. In this manner, the Small Cluster model creates an open-set label noise scenario \cite{wang2018iterative} where some images that do not belong to any other existing class in the training set. 

\begin{table*}[t!]
\centering
\caption{Precision@1 / MAP@R (\%) on CARS dataset with synthetic label noise.
}\label{tab:cars_result}
\resizebox{1.0\linewidth}{!}{ \begin{tabular}{@{}lrrrrrrrrrrr@{}}\toprule
     & \multicolumn{4}{c}{Symmetric Noise}                && \multicolumn{4}{c}{Small Cluster Noise}             &                   \\ \cmidrule{2-5} \cmidrule{7-10}
Noisy Label Rate & 10\%           & 20\%           & 50\% & 70\% && 10\%           & 25\%           & 50\% &75\%           \\ \midrule
\multicolumn{10}{l}{\textit{Algorithms for image classification under label noise}}                                                \\
Co-teaching \cite{han2018co}                                          & 75.25 / 20.26 & 71.92 / 18.77 & 65.82 / 14.69 & 58.78 / 10.49 && 77.53 / 23.39 & 74.98 / 21.75 & 73.46 / 18.76 & 63.29 / 12.52 \\
Co-teaching+ \cite{yu2019does}                                        & 72.97 / 19.81 & 73.66 / 19.61 & 63.47 / 13.97 & 58.18 / 10.29 && 78.17 / 23.59 & 74.90 / 21.72 & 72.69 / 18.70 & 62.45 / 12.09 \\
Co-teaching \cite{han2018co} w/ Temperature \cite{zhaiclassification} & 80.88 / 24.82 & 81.13 / 24.33 & 73.96 / 18.89 & 59.20 / 9.87  && 82.09 / 26.94 & 80.98 / 23.50 & 74.78 / 19.76 & 56.00 / 8.11  \\
F-correction \cite{patrini2017making}                                 & 78.07 / 22.19 & 76.06 / 17.69 & 71.22 / 20.91 & 56.66 / 10.02 &&   N/A        &    N/A      &   N/A       &       N/A    \\ \midrule
\multicolumn{10}{l}{\textit{DML with Proxy-based Losses}}                                                                          \\
FastAP \cite{cakir2019deep}                                           & 73.62 / 23.11 & 71.54 / 21.74 & 63.24 / 16.18 & 50.00 / 6.93  && 72.54 / 22.54 & 68.96 / 20.00 & 62.67 / 15.94 & 43.22 / 6.41  \\
nSoftmax \cite{zhaiclassification}                                    & 79.40 / 21.90 & 76.60 / 19.03 & 67.02 / 12.99 & 58.28 / 9.03  && 80.48 / 25.48 & 77.11 / 19.79 & 70.09 / 15.13 & 60.90 / 9.46  \\
ProxyNCA \cite{movshovitz2017no}                                      & 78.88 / 26.41 & 78.83 / 25.38 & 68.81 / 17.29 & 60.56 / 10.86 && 77.65 / 22.91 & 77.95 / 23.83 & 70.97 / 18.22 & 60.56 / 11.28 \\
SoftTriple \cite{qian2019softtriple}                                  & 82.51 / 23.84 & 79.57 / 20.48 & 67.12 / 12.40 & 58.01 / 8.72  && 84.50 / 26.53 & 80.45 / 22.25 & 72.99 / 17.04 & 61.57 / 10.78 \\
SoftTriple + PRISM (AvgSim) (Ours)                                    & 83.30 / 26.49 & 83.60 / 26.04 & 75.54 / 19.37 & 63.58 / 11.29 && 85.95 / 28.59 & \textbf{84.26} / 25.58 & 76.18 / 19.70 & 61.27 / 12.06 \\
SoftTriple + PRISM (ProxySim) (Ours)                                  & 82.22 / 24.24 & 82.93 / 25.29 & 74.36 / 18.78 & 57.27 / 9.53  && 84.85 / 28.20 & 82.02 / 25.78 & 76.67 / \textbf{20.55} & \textbf{65.30 / 13.45}  \\
SoftTriple + PRISM (vMF-Sim) (Ours)                                       & \textbf{83.62 / 28.93} & \textbf{85.04 / 28.91} & \textbf{77.92 / 23.06} & \textbf{65.94 / 13.57} && \textbf{86.67 / 29.29} & 83.67 / \textbf{26.61} & \textbf{76.82} / 20.00 & 60.36 / 11.82 \\ \midrule
\multicolumn{10}{l}{\textit{DML with Pair-based Losses}}                                                                           \\
MS \cite{wang2019multi}                                               & 77.90 / 21.93 & 72.59 / 16.43 & 47.69 / 5.90  & 45.53 / 5.31  && 77.75 / 22.52 & 71.41 / 16.25 & 51.82 / 7.74  & 45.70 / 5.33  \\
Circle \cite{sun2020circle}                                           & 76.94 / 20.83 & 61.59 / 10.70 & 45.21 / 5.24  & 44.74 / 5.22  && 77.68 / 22.34 & 62.62 / 11.90 & 45.26 / 5.28  & 45.95 / 5.29  \\
Contrastive Loss \cite{chopra2005learning}                            & 76.74 / 23.00 & 76.60 / 21.17 & 46.51 / 5.30  & 41.09 / 4.90  && 75.49 / 20.61 & 71.22 / 18.08 & 52.70 / 8.23  & 41.70 / 4.97  \\
Memory Contrastive Loss (MCL) \cite{wang2020cross}                    & 75.96 / 19.20 & 75.12 / 17.23 & 45.73 / 3.40  & 46.02 / 5.30  && 80.13 / 23.95 & 67.29 / 13.42 & 49.26 / 6.73  & 45.80 / 5.27  \\
MCL + PRISM (AvgSim) (Ours)                                           & 81.51 / 26.55 & \textbf{82.91} / 26.39 & 75.42 / 20.63 & 65.50 / 12.06 && \textbf{83.57} / 27.50 & 80.94 / 24.60 & 75.27 / 20.03 & \textbf{60.12} / 11.7  \\
MCL + PRISM (vMF-Sim) (Ours)                                              & \textbf{82.86 / 27.60} & 82.74 / \textbf{27.21} & \textbf{77.90 / 23.98} & \textbf{68.93 / 16.56} && 83.18 / \textbf{27.60} & \textbf{81.56 / 25.37} & \textbf{75.88 / 21.11} & \textbf{60.12 / 11.74} \\ \bottomrule
\end{tabular} }
\end{table*}

\subsection{Baseline Techniques}\label{sec:setting}
We compare PRISM against 12 baseline approaches, including four designed for noise-resistant classification and eight designed for deep metric learning. The four noise-resistant classification baselines include Co-teaching \cite{han2018co}, Co-teaching+ \cite{yu2019does}, Co-teaching with Temperature \cite{zhaiclassification}, and F-correction \cite{patrini2017making}. Among these, the F-correction method assumes closed-world noise and can only be used under symmetric noise. 

Among the deep learning metric baselines, we adopt four with proxy-based losses (SoftTriple \cite{qian2019softtriple}, FastAp \cite{cakir2019deep}, nSoftmax \cite{zhaiclassification} and proxyNCA \cite{movshovitz2017no}), and four using pair-based losses (MS loss \cite{wang2019multi}, circle loss \cite{sun2020circle}, contrastive loss \cite{chopra2005learning} and memory contrastive loss (MCL) \cite{wang2020cross}). All the DML baselines assume that data samples are clean.
We train the classification baselines using cross-entropy, use the $l_2$-normalized features from the layer before the final linear classifier when retrieving images during inference.

We use the official implementation for Co-teaching, Co-teaching+ and Memory Contrastive Loss \cite{wang2020cross}, and the PyTorch Metric Learning library \cite{musgrave2020pytorch} for the other DML algorithms. 
For Co-teaching and Co-teaching+, we use the learning rate (LR) scheduler given in their code, while for others (including PRISM), the cosine LR decay \cite{loshchilov2016sgdr} is used. We set the batch size to 64 for experiments on all datasets and all models. During training, the input images are first resized to 256x256, then randomly cropped to 224x224. A horizontal flip is performed on the training data with a probability of 0.5. The validation and testing images are resized to 224x224 without data augmentation.


Following \cite{wang2020cross}, when comparing performance on CARS, CUB and CARS-98N, we use BN-inception \cite{ioffe2015batch} as the backbone CNN model for all algorithms. The dimension of the output feature is set as 512, the same as in \cite{wang2020cross}. No other tricks (e.g., freezing BN layers) are used during the experiments. For SOP and Food-101N, we use ResNet-50 \cite{he2016deep} with a 128-dimensional output. 

Evaluation and testing is based on the ranked list of the nearest neighbors for the evaluation/test images. Specifically, we use Precision@1 (P@1) and Mean Average Precision@R (MAP@R) \cite{musgrave2020metric} as the evaluation metrics. The test and validation sets are noise-free, as the purpose of the experiments is to evaluate the algorithms in the presence of noisy labels in the training data.

\subsection{Results}\label{sec:results}

\begin{table*}[t!]
\centering
\caption{Precision@1 / MAP@R (\%) on CUB dataset with synthetic label noise. 
}\label{tab:cub_result}
\resizebox{1.0\linewidth}{!}{ \begin{tabular}{@{}lrrrrrrrrrrr@{}}\toprule
     & \multicolumn{4}{c}{Symmetric Noise}                && \multicolumn{4}{c}{Small Cluster Noise}             &                   \\ \cmidrule{2-5} \cmidrule{7-10}
Noisy Label Rate & 10\%           & 20\%           & 50\% & 70\% && 10\%           & 25\%           & 50\% &75\%           \\ \midrule
\multicolumn{10}{l}{\textit{Algorithms for image classification under label noise}}                                                \\
Co-teaching \cite{han2018co}                                          & 60.94 / 19.60 & 55.57 / 18.72 & 52.00 / 14.84 & 48.88 / 12.85 && 57.24 / 19.45 & 54.25 / 17.92 & 53.94 / 15.84 & 51.19 / 15.30 \\
Co-teaching+ \cite{yu2019does}                                        & 60.46 / 19.71 & 55.84 / 17.72 & 52.00 / 13.88 & 49.59 / 13.92 && 58.93 / 19.56 & 55.10 / 17.80 & 53.26 / 15.45 & 51.49 / 15.22 \\
Co-teaching \cite{han2018co} w/ Temperature \cite{zhaiclassification} & 60.43 / 20.59 & 59.82 / 19.97 & 56.59 / 18.41 & 50.85 / 14.24 && 60.67 / 20.11 & 59.92 / 19.77 & 56.39 / 16.74 & 45.45 / 11.97 \\
F-correction \cite{patrini2017making}                                 & 56.88 / 19.02 & 57.62 / 18.67 & 51.01 / 14.98 & 46.12 / 11.72 &&      N/A       &   N/A          &     N/A     &       N/A     \\ \midrule
\multicolumn{10}{l}{\textit{DML with Proxy-based Losses}}                                                                          \\
FastAP \cite{cakir2019deep}                                           & 59.81 / 22.89 & 59.20 / 21.90 & 56.07 / 19.73 & 48.28 / 15.56 && 59.51 / 22.02 & 57.18 / 20.45 & 52.66 / 17.80 & 48.08 / 14.16 \\
nSoftmax \cite{zhaiclassification}                                    & 55.29 / 17.46 & 53.37 / 15.86 & 52.16 / 14.11 & 51.28 / 12.52 && 59.84 / 19.69 & 56.17 / 17.47 & 52.29 / 14.70 & 51.62 / 13.78 \\
ProxyNCA \cite{movshovitz2017no}                                      & 58.50 / 20.89 & 58.02 / 20.81 & 55.19 / 17.14 & 49.43 / 15.10 && 59.44 / 21.11 & 58.29 / 20.10 & 57.11 / 19.25 & 50.37 / 14.47 \\
SoftTriple \cite{qian2019softtriple}                                  & 63.72 / 21.70 & 60.08 / 18.95 & 51.18 / 13.93 & 47.91 / 12.42 && 64.94 / 23.81 & 59.51 / 20.47 & 56.37 / 17.36 & 49.16 / 12.95 \\
SoftTriple + PRISM (AvgSim) (Ours)                                    & 63.05 / 22.47 & 64.03 / 22.07 & 59.10 / 19.34 & 51.15 / 14.08 && 65.51 / 24.18 & 63.01 / 22.55 & 57.55 / 19.45 & 51.48 / 14.89 \\
SoftTriple + PRISM (ProxySim) (Ours)                                  & 63.76 / 21.73 & 62.37 / 21.32 & 56.54 / 17.11 & 51.79 / 13.89 && 65.27 / 23.82 & 62.44 / 21.52 & 58.63 / 19.75 & \textbf{53.44 / 16.10} \\
SoftTriple + PRISM (vMF-Sim) (Ours)                                       & \textbf{66.52 / 24.01} & \textbf{64.13 / 23.71} & \textbf{61.09 / 20.43} & \textbf{54.92/ 16.48} && \textbf{65.98 / 24.53} & \textbf{64.94 / 23.16} & \textbf{59.07 / 20.40} & 53.17 / 15.76 \\ \midrule
\multicolumn{10}{l}{\textit{DML with Pair-based Losses}}                                                                           \\
MS \cite{wang2019multi}                                               & 61.26 / 21.41 & 58.63 / 19.52 & 43.26 / 9.69  & 36.24 / 6.80  && 62.61 / 23.09 & 56.98 / 18.57 & 46.19 / 10.95 & 40.26 / 7.83  \\
Circle \cite{sun2020circle}                                               & 51.82 / 16.14 & 49.56 / 12.66 & 35.10 / 6.40  & 34.39 / 6.05  && 57.59 / 18.14 & 47.98 / 12.96 & 39.89 / 7.78  & 39.95 / 7.57  \\
Contrastive Loss \cite{chopra2005learning}                            & 56.81 / 18.45 & 54.99 / 21.22 & 44.67 / 9.89  & 42.58 / 8.94  && 56.34 / 16.28 & 53.51 / 16.13 & 50.51 / 12.93 & 42.38 / 9.18  \\
Memory Contrastive Loss (MCL) \cite{wang2020cross}                    & 61.70 / 20.34 & 61.94 / 15.58 & 41.98 / 8.68  & 41.98 / 8.79  && 64.50 / 22.83 & 58.87 / 18.30 & 47.13 / 10.51 & 35.13 / 6.26  \\
MCL + PRISM (AvgSim) (Ours)                                           & 65.21 / 23.65 & 63.86 / 22.38 & 60.72 / 21.10 & 51.89 / 14.74 && \textbf{65.71} / 23.85 & 63.65 / 22.37 & 58.63 / 19.77 & \textbf{51.42} / 14.37 \\
MCL + PRISM (vMF-Sim) (Ours)                                              & \textbf{66.62 / 26.33} & \textbf{66.01 / 25.04} & \textbf{63.59 / 23.24} & \textbf{57.05 / 19.14} && 65.04 / \textbf{25.02} & \textbf{64.13 / 23.00} & \textbf{58.77 / 19.88} & 49.90 / \textbf{15.02} \\ \bottomrule
\end{tabular} }
\end{table*}

Table~\ref{tab:cars_result} shows the performance of the algorithms on CARS. As we expect, higher level noise always leads to reduced performance. PRISM achieves the best performance and suffers the least performance degradation when the noise level increases. Algorithms for image classification under label noise performs generally worse than the deep metric learning methods, which highlights the need for developing noise filtering algorithm for DML. 

In proxy-based methods, PRISM improves the performance of SoftTriple under all noise models and levels. No matter what $P_{\text{clean}}(i)$ is chosen, SoftTriple+PRISM performs better than all the compared proxy-based methods. vMF-Sim stands out as the best performing method when $R<70\%$, indicating that it is beneficial to explicitly model class tightness. AvgSim is better than ProxySim for most cases. Under high levels of noise ($R \geq 70\%$), with symmetric noise, vMF-Sim still produces the best results. However, with 75\% Small Cluster noise, ProxySim performs better than vMF-Sim. This is because when noise level is high, the parameter estimates become inaccurate \cite{kato2016robust}. Neverthless, high noise levels with $R \geq 75\%$ is rare in practice.  

Comparing the pair-based methods with the proxy-based methods, the performance of algorithms without noise filtering is generally lower, which indicates pair-based losses are more vulnerable to the label noise. It is because $R\%$ label noise results in clean pairs accounting for only $(1-R)^2\%$. Such a large-scale incorrect supervision negatively impact model performance. PRISM significantly boosts the performance of pair-based loss. As the noise level increases, the performance gap between PRISM and the other pair-based methods becomes larger. vMF-Sim outperforms AvgSim.

Tables~\ref{tab:cub_result}-\ref{tab:real_result} show results under the CUB, SOP, and the two real-world datasets, CARS-98N and Food-101N. 
Under CUB, PRISM (vMF-Sim) achieves the best performance for all situations except 75\% Small Cluster noise.
Under SOP, for almost all cases, PRISM (AvgSim) achieves the best performance. Unlike other datasets, the performance of vMF-Sim is generally worse than AvgSim on SOP. We investigate the reason behind this phenomenon in Section~\ref{sec:discuss}.
Under the real-world datasets CARS-98N and Food-101N, PRISM also achieves the best performance. SoftTriple + ProxySim achieves the best performance under Food-101N. 

\begin{table*}[t!]
\centering
\caption{Precision@1 (\%) on SOP dataset with synthetic label noise. 
}\label{tab:sop_result}
 \begin{tabular}{@{}lrrrrrrrrrrr@{}}\toprule
     & \multicolumn{4}{c}{Symmetric Noise}                && \multicolumn{4}{c}{Small Cluster Noise}             &                   \\ \cmidrule{2-5} \cmidrule{7-10}
Noisy Label Rate & 10\%           & 20\%           & 50\% & 70\% && 10\%           & 25\%           & 50\% &75\%           \\ \midrule
\multicolumn{10}{l}{\textit{Algorithms for image classification under label noise}}                                                \\
Co-teaching \cite{han2018co}                                          & 55.93 & 55.16 & 56.07 & 53.70 && 54.54 & 55.20 & 54.79 & 54.71 \\
Co-teaching+ \cite{yu2019does}                                        & 54.31 & 54.87 & 54.95 & 53.70 && 51.04 & 55.26 & 54.95 & 54.99 \\
Co-teaching \cite{han2018co} w/ Temperature \cite{zhaiclassification} & 67.36 & 74.30 & 51.09 & 49.25 && 73.14 & 71.00 & 66.26 & 62.06 \\
F-correction \cite{patrini2017making}                                 & 54.41 & 55.02 & 54.93 & 53.10 &&   N/A  &   N/A   &   N/A   &   N/A   \\ \midrule
\multicolumn{10}{l}{\textit{DML with Proxy-based Losses}}                                                                          \\
FastAP \cite{cakir2019deep}                                           & 73.71 & 72.66 & 68.78 & 63.46 && 73.78 & 72.56 & 70.25 & 64.02 \\
nSoftmax \cite{zhaiclassification}                                    & 77.38 & 74.21 & 59.82 & 44.35 && 75.98 & 73.55 & 69.53 & 63.34 \\
ProxyNCA \cite{movshovitz2017no}                                      & 75.08 & 73.86 & 67.53 & 61.27 && 74.86 & 74.00 & 67.31 & 61.47 \\
SoftTriple \cite{qian2019softtriple}                                  & 74.18 & 76.50 & 66.64 & 63.14 && 79.38 & 75.98 & 74.54 & 63.52 \\
SoftTriple + PRISM (AvgSim) (Ours)                                    & \textbf{78.90} & \textbf{78.01} & 70.79 & \textbf{66.97} && \textbf{79.21} & 76.65 & \textbf{75.71} & 67.11 \\
SoftTriple + PRISM (ProxySim) (Ours)                                  & 74.72 & 74.43 & 67.67 & 64.32 && 75.09 & 76.82 & 74.52 & \textbf{68.62} \\
SoftTriple + PRISM (vMF-Sim) (Ours)                                       & 76.10 & 73.22 & \textbf{71.34} & 65.30 && 73.27 & \textbf{78.35} & 75.31 & 66.35 \\\midrule
\multicolumn{10}{l}{\textit{DML with Pair-based Losses}}                                                                           \\
MS \cite{wang2019multi}                                               & 76.00 & 74.28 & 60.82 & 58.22 && 76.66 & 74.21 & 65.34 & 59.84 \\
Circle \cite{sun2020circle}                                           & 76.91 & 75.10 & 60.23 & 55.22 && 77.06 & 75.27 & 47.58 & 51.37 \\
Contrastive Loss \cite{chopra2005learning}                            & 73.81 & 73.68 & 69.46 & 60.38 && 73.80 & 73.05 & 71.85 & 59.96 \\
Memory Contrastive Loss (MCL) \cite{wang2020cross}                    & 82.21 & 80.34 & 71.20 & 58.68 && 82.29 & 79.44 & 71.63 & 54.52 \\
MCL + PRISM (AvgSim) (Ours)                                           & \textbf{83.08} & \textbf{82.91} & \textbf{76.45} & \textbf{66.22} && \textbf{83.30} & \textbf{82.04} & 76.35 & 68.21 \\
MCL + PRISM (vMF-Sim) (Ours)                                              & 81.52 & 80.45 & 76.18 & 65.32 && 80.76 & 80.40 & \textbf{77.44} & \textbf{70.08} \\ \bottomrule
\end{tabular}
\end{table*}

\begin{table}[t!]
\centering
\caption{Precision@1 / MAP@R (\%) on Real-world noise dataset, CARS-98N and Food-101N.
}\label{tab:real_result}
\resizebox{1.0\linewidth}{!}{ \begin{tabular}{@{}lrr@{}}\toprule
                                                                      & CARS-98N      & Food-101N     \\ \midrule
\multicolumn{3}{l}{\textit{Algorithms for image classification under label noise}}                                                \\
Co-teaching \cite{han2018co}                                          & 60.49 / 11.20 & 61.77 / 18.17 \\
Co-teaching+ \cite{yu2019does}                                        & 60.49 / 10.80 & 59.14 / 16.99 \\
Co-teaching \cite{han2018co} w/ Temperature \cite{zhaiclassification} & 69.02 / 15.18 & 68.49 / 22.30 \\ \midrule
\multicolumn{3}{l}{\textit{DML with Proxy-based Losses}}                                                                          \\
FastAP \cite{cakir2019deep}                                           & 52.04 / 9.88  & 52.58 / 12.46 \\
nSoftmax \cite{zhaiclassification}                                    & 66.97 / 13.64 & 68.15 / 19.61 \\
ProxyNCA \cite{movshovitz2017no}                                      & 59.48 / 12.59 & 59.50 / 16.70 \\
Softtriple \cite{qian2019softtriple}                                  & 70.75 / 16.25 & 69.64 / 21.18 \\
SoftTriple + PRISM (AvgSim) (Ours)                                    & 71.83 / \textbf{16.63} & 69.33 / 20.91 \\
SoftTriple + PRISM (ProxySim) (Ours)                                  & 71.19 / 15.94 & \textbf{70.59 / 22.32} \\
SoftTriple + PRISM (vMF-Sim) (Ours)                                       & \textbf{72.20} / 16.16 & 69.92 / 22.13 \\ \midrule
\multicolumn{3}{l}{\textit{DML with Pair-based Losses}}                                                                           \\
MS \cite{wang2019multi}                                               & 59.21 / 10.05 & 58.86 / 13.25 \\
Circle \cite{sun2020circle}                                           & 41.04 / 4.97  & 50.26 / 9.77  \\
Contrastive Loss \cite{chopra2005learning}                            & 59.77 / 10.61 & 57.19 / 12.17 \\
Memory Contrastive Loss (MCL) \cite{wang2020cross}                    & 57.27 / 8.28  & \textbf{59.58} / 14.27 \\
MCL + PRISM (AvgSim) (Ours)                                           & 63.70 / 11.28 & 58.89 / 14.01 \\
MCL + PRISM (vMF-Sim) (Ours)                                              & \textbf{64.15 / 11.68} & 59.19 / \textbf{14.30} \\ \bottomrule
\end{tabular}}
\end{table}

\begin{table}[]
\caption{Training time (seconds) required with and without PRISM (AvgSim, sTRM) for 5,000 iterations on SOP dataset. The time recording starts when the memory bank is completely filled at iteration 3,000.}\label{tab:SOP_time}
\resizebox{1\columnwidth}{!}{
\begin{tabular}{@{}ll@{}}\toprule
Algorithm              & Training Time (Seconds) \\ \midrule
Memory Contrastive Loss (MCL)       & 1,679.22        \\ 
MCL + PRISM (AvgSim) without centers         & 12,294.76       \\ 
MCL + PRISM (AvgSim) with centers            & 1,777.38 (+5.8\%)        \\ \midrule
SoftTriple                         & 1,685.47        \\ 
SoftTriple + PRISM (AvgSim) with centers    & 1,767.97 (+4.9\%)        \\ \bottomrule
\end{tabular}
}

\end{table}

\begin{table}[]

\caption{Training time (seconds) on the CUB and CARS datasets. The setting is the same as Table~\ref{tab:SOP_time}.}\label{tab:cub_time}
\resizebox{1\columnwidth}{!}{ 
\begin{tabular}{llll}\toprule
Algorithm                     & CARS               & CUB                \\\midrule
MCL                           & 1,153.18             & 1,149.74           \\
MCL+ PRISM (AvgSim)           & 1,253.99 (+8.7\%) & 1,266.04 (+10.1\%)  \\
MCL+ PRISM (vMF-Sim)              & 1,604.06 (+39.1\%) & 1,602.64 (+39.4\%) \\\midrule
SoftTriple                    & 1,146.38           & 1,134.76           \\
SoftTriple+ PRISM (AvgSim)    & 1,250.15 (+9.1\%)  & 1,271.18 (+12.0\%) \\
SoftTriple+ PRISM (ProxySim)  & 1,200.84 (+4.8\%)  & 1,161.77 (+2.4\%)\\
SoftTriple+ PRISM (vMF-Sim)       & 1,636.11 (+42.7\%) & 1,641.63 (+44.7\%) \\\bottomrule
\end{tabular}}

\end{table}

\vspace{0.1in}
\noindent \textbf{Training Time.} To study the time efficiency of PRISM, we report the time required for 5,000 training iterations in Tables~\ref{tab:SOP_time} and \ref{tab:cub_time}. 
PRISM without centers (Eq.~\eqref{equ:pi}) requires the longest training time because it needs to calculate average similarities of all classes for each minibatch data. However, by maintaining center vectors to identify noise (Eq.~\eqref{equ:pi2}), the required training time decreases significantly. PRISM only incurs 6\% more training time than that of Memory Contrastive Loss on SOP. Similar results can be observed when we change the loss function to SoftTriple \cite{qian2019softtriple}. Under CARS and CUB dataset, ProxySim is the fastest method to filter noise among AvgSim, vMF-Sim and ProxySim, and only incurs up to 4.8\% more training time. vMF-Sim incurs about 39\% additional training time due to the slow calculation of $\log I_{\frac{D}{2}-1}(\cdot)$. Improving the speed of the von Mises-Fisher distribution estimation may be an interesting direction for the future. Overall, the results show that PRISM is an efficient label noise filtering approach.

\subsection{Analysis}\label{sec:discuss}

\begin{figure}
    \centering

    \includegraphics[width=0.8\columnwidth]{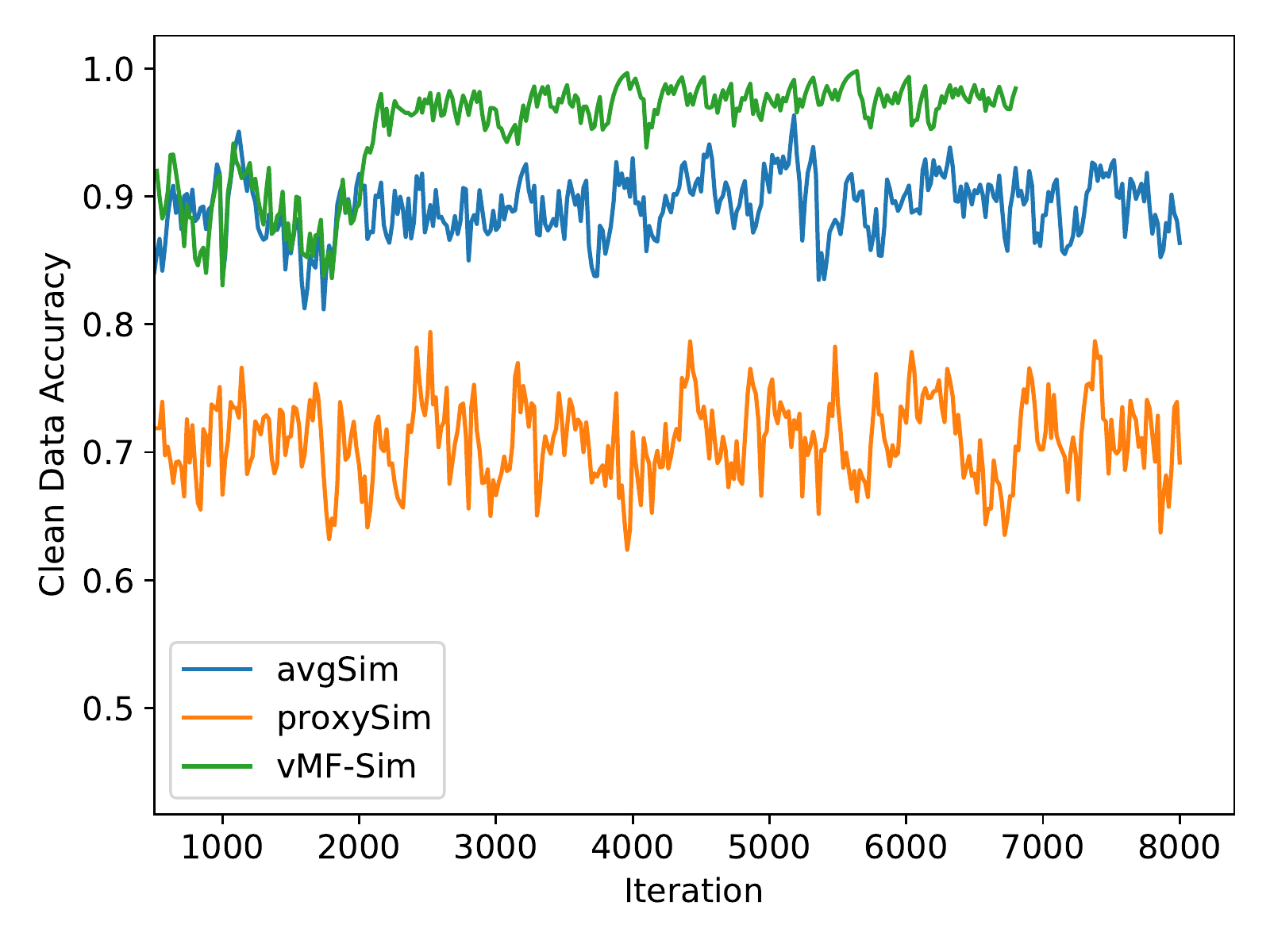}
        \caption{The accuracy of noise filtering vs number of iterations. A exponential moving average with $\text{weight}=0.7$ is performed to smooth the curve. Models are trained on CUB with 50\% Symmetric Noise with SoftTriple loss. In the first 1,500 iterations of vMF-Sim training, AvgSim is used (i.e. $I_v=1500$). The vMF-Sim training is early stopped at the 6,800 iteration.}
    
    \label{fig:cleanness}
\end{figure}

\vspace{0.1in}\noindent\textbf{Dynamics of the cleanness probability}. As discussed earlier, a crucial consideration of DML under noise is to bootstrap the identification of noise data. That is, we need to learn good representations that identify noise even when noise is present in the training data. To investigate whether the algorithms achieve this purpose, we compute the accuracy of noise identification as training progresses and show the results in Figure~\ref{fig:cleanness}. Since we have ground-truth labels for synthetic noise datasets, we measure the portion of ground-truth clean data from all data considered clean. 

We observe that vMF-Sim achieves the best noise filtering accuracy. After 2,500 iterations, vMF-Sim achieves close to 100\% accuracy. The ProxySim method is worse than AvgSim in most iterations. This is consistent with superior performance of vMF-Sim as reported in Table~\ref{tab:cub_result}.

\begin{figure}[]
    \centering
    \subfigure[AvgSim, the 2,000th training iteration]{\includegraphics[width=0.48\columnwidth]{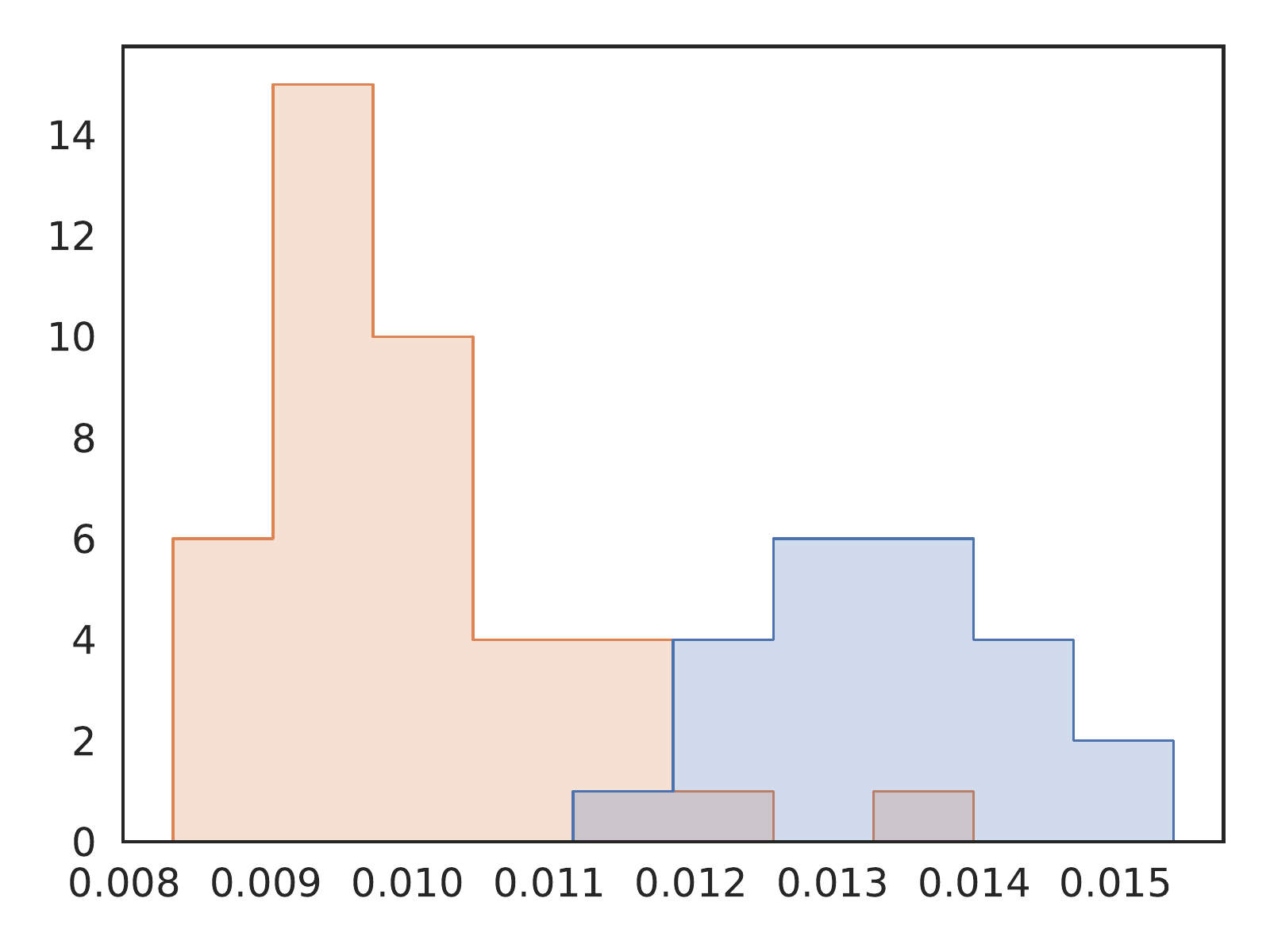}}
    \subfigure[AvgSim, the 3,000th training iteration]{\includegraphics[width=0.48\columnwidth]{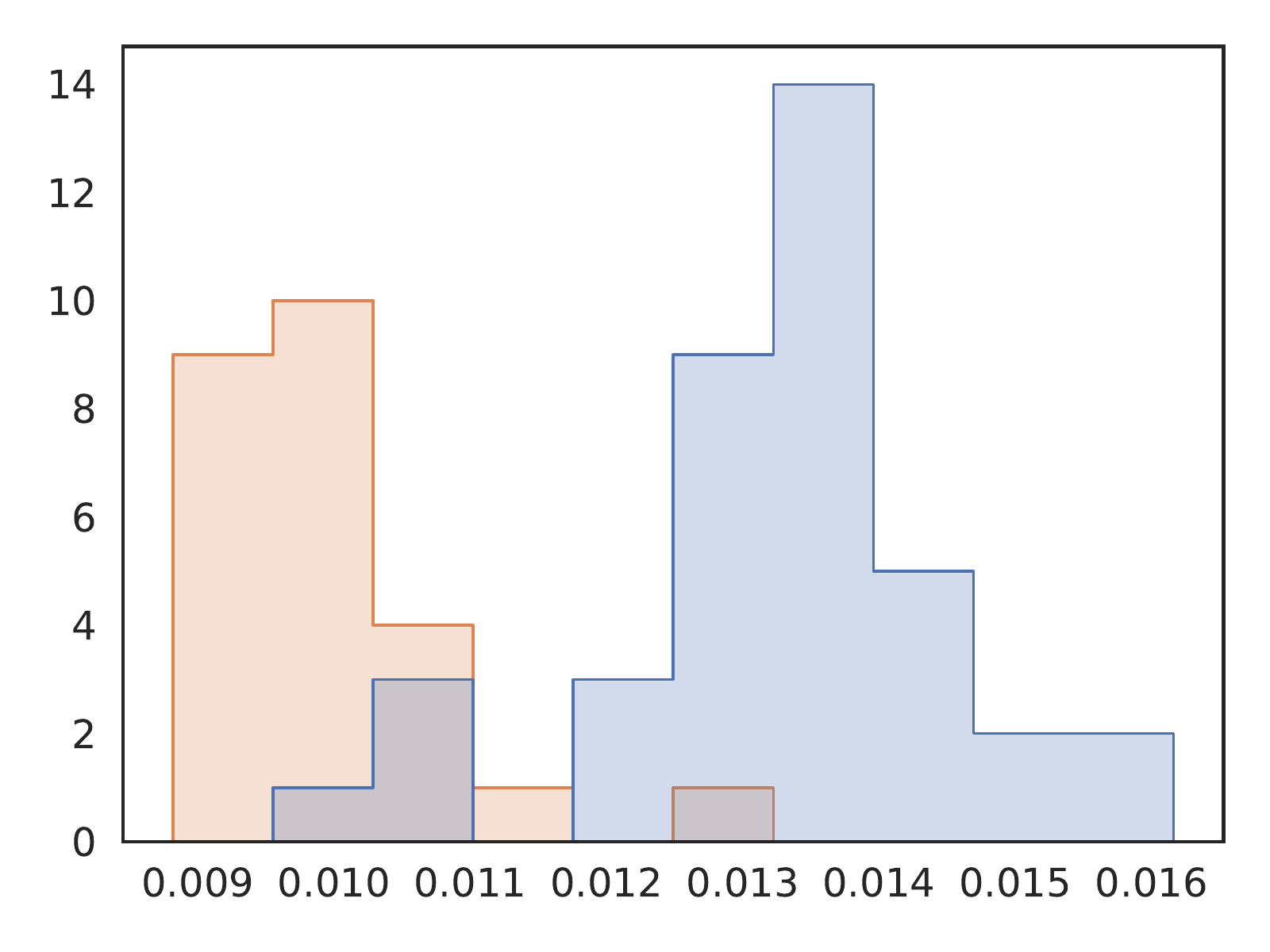}}
    \subfigure[ProxySim, the 2,000th training iteration]{\includegraphics[width=0.48\columnwidth]{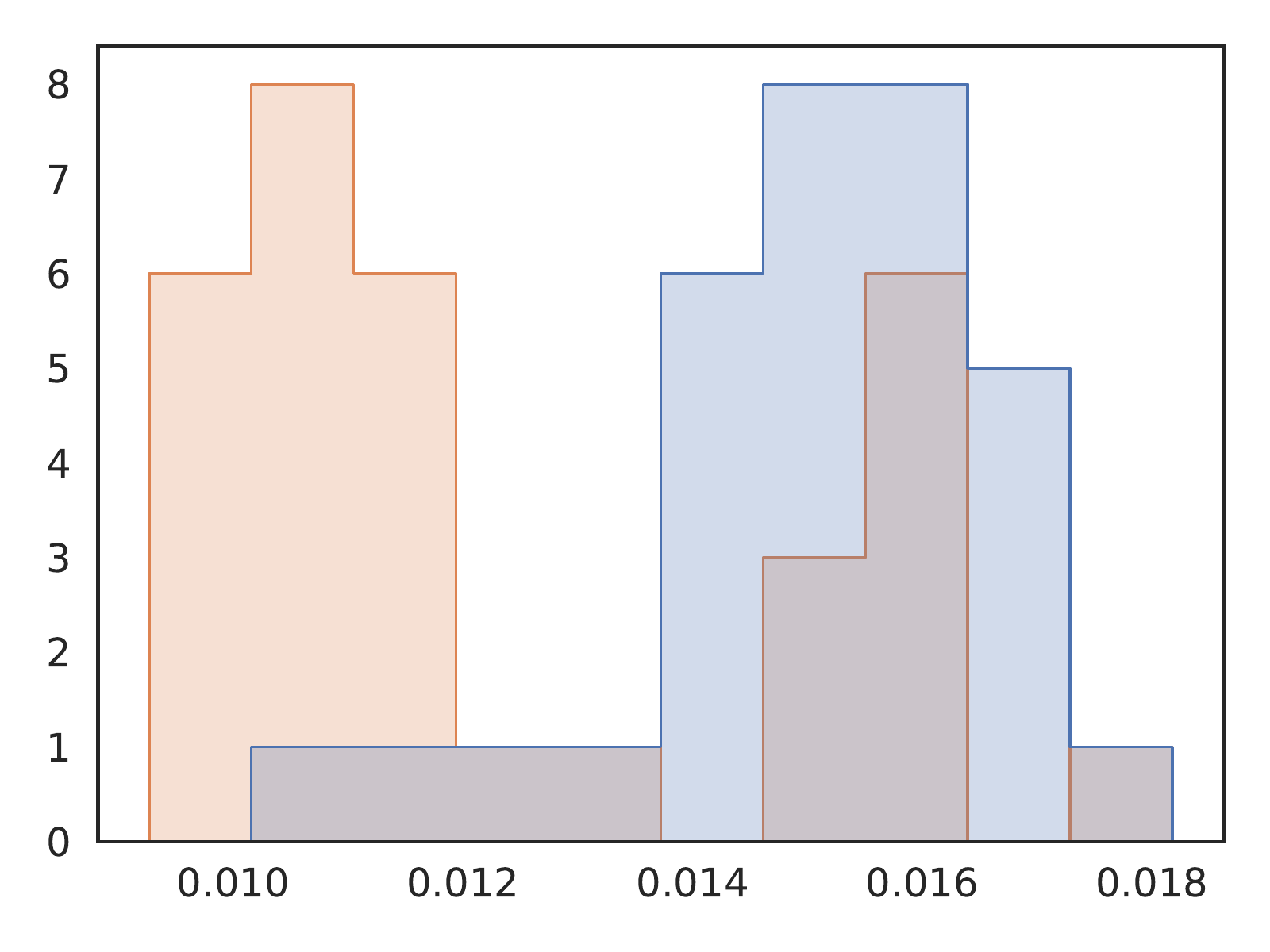}}
    \subfigure[ProxySim, the 3,000th training iteration]{\includegraphics[width=0.48\columnwidth]{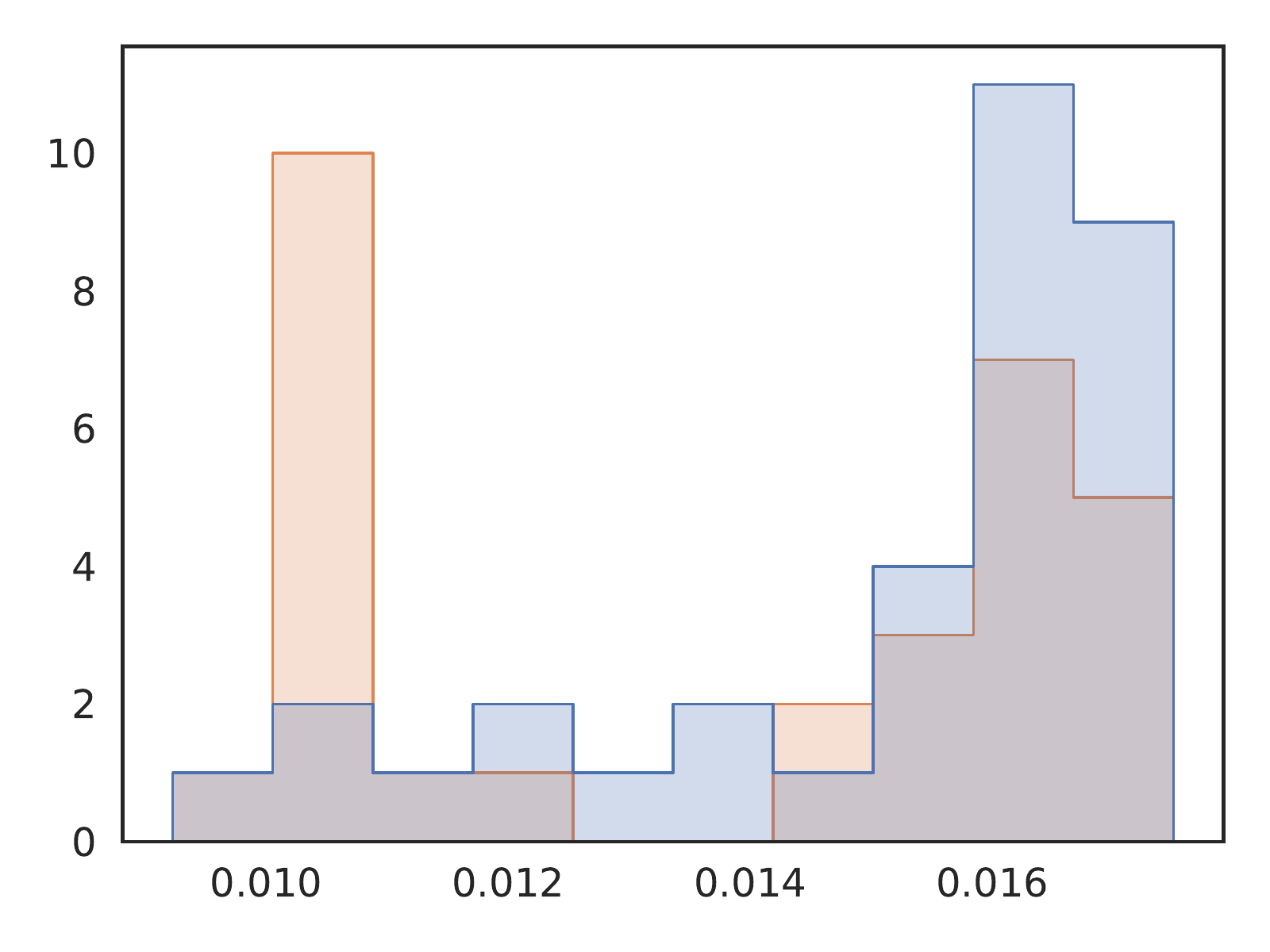}}
    \subfigure[vMF-Sim, the 2,000th training iteration]{\includegraphics[width=0.48\columnwidth]{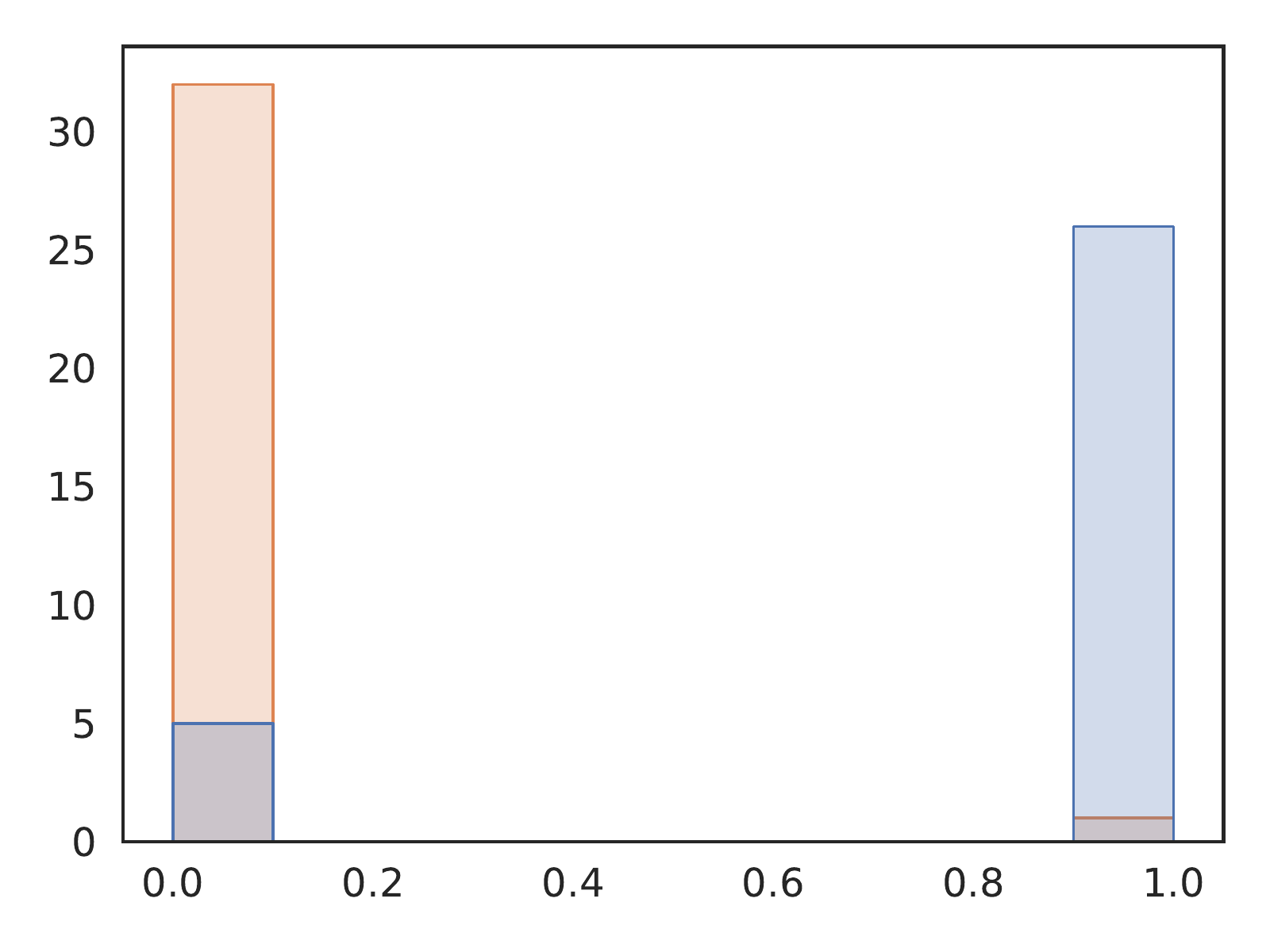}}
    \subfigure[vMF-Sim, the 3,000th training iteration]{\includegraphics[width=0.48\columnwidth]{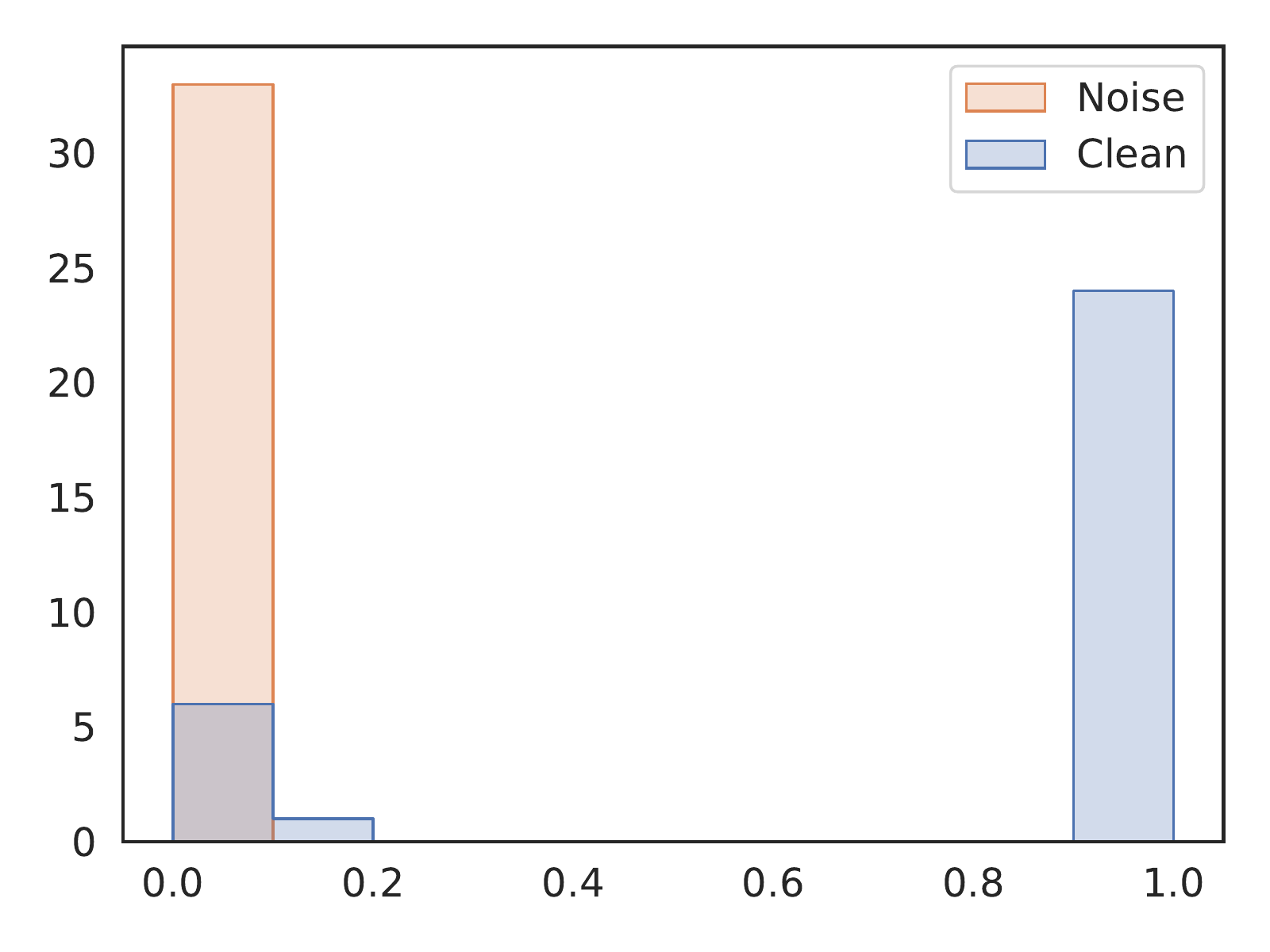}}
    \caption{Histogram of $P_{\text{clean}}(i)$ from different calculation methods at different iterations. The x-axis shows the value of $P_{\text{clean}}(i)$ and the y-axis shows the count of occurrence of the corresponding $P_{\text{clean}}(i)$ value. The dataset used is CUB with 50\% symmetric noise.}
    \label{fig:pclean_histogram}
\end{figure}

\noindent\textbf{vMF-Sim separates the clean and noisy data better than AvgSim and ProxySim}. We show the distributions of $P_{\text{clean}}(i)$ for the three techniques in Figure~\ref{fig:pclean_histogram}. For AvgSim and ProxySim, the distributions between clean and noisy data have significant overlap, whereas for vMF-Sim, most noisy data have the probability $P_{\text{clean}}(i)$ close to 0 and most clean data have $P_{\text{clean}}(i)$ close to 1.


\begin{figure}[t!]
      \centering
      
      \includegraphics[width=0.7\columnwidth]{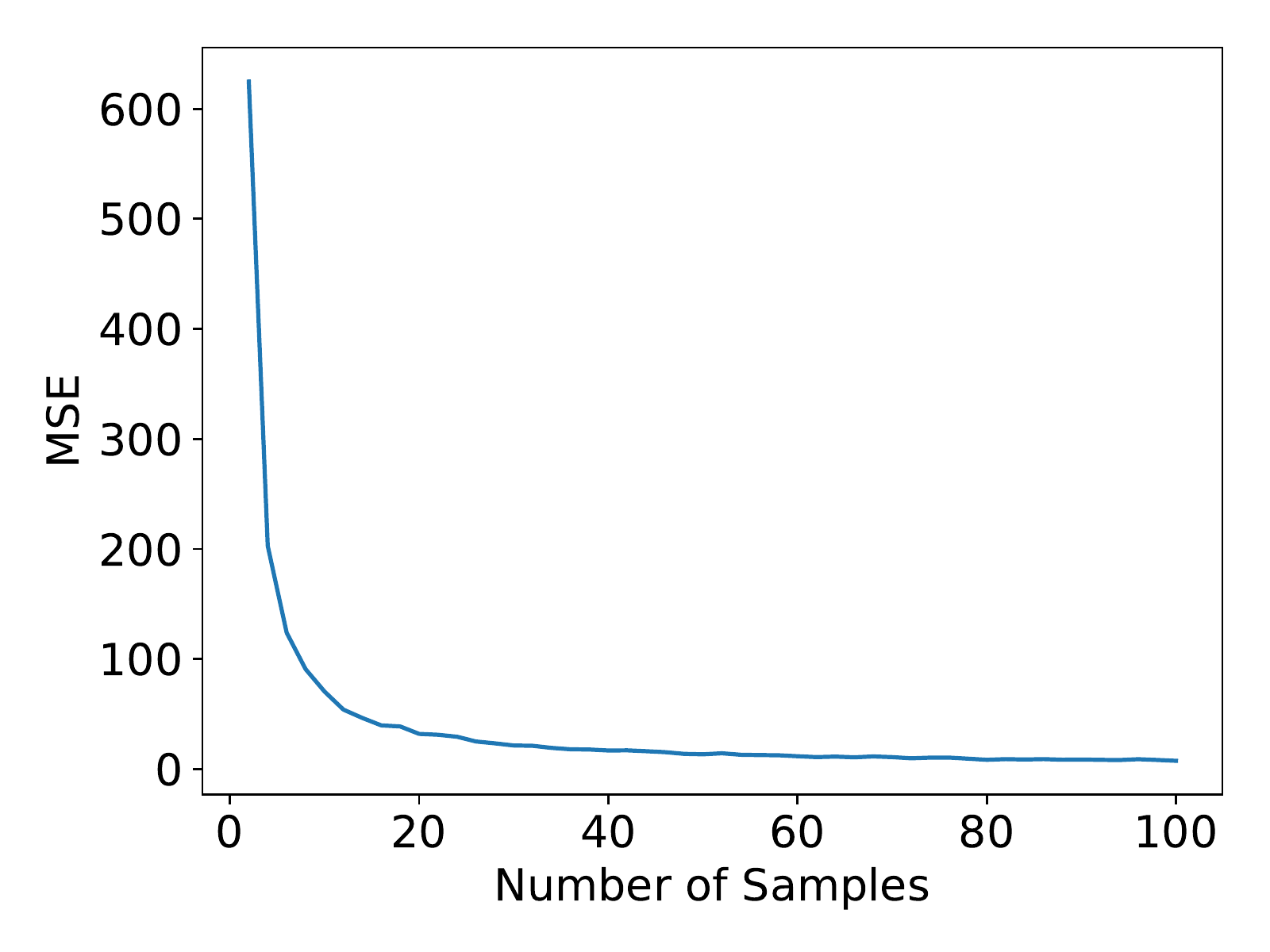}
      \caption{Number of samples vs. Mean Squared Error (MSE) of the estimated $\kappa$. We sample data from a vMF distribution with a random $\mu$ and $\kappa=537$ on $\mathbb{S}^{127}$, and use Eq.~\ref{equ:vMF_kappa} to estimate $\kappa$. Every result is the average over 200 independent trials.}\label{fig:vMF_mse}
\end{figure}

\vspace{0.1in}\noindent\textbf{vMF-Sim under SOP}. Although vMF-Sim performs better than AvgSim on most datasets, on the SOP dataset, vMF-Sim underperforms AvgSim. We hypothesize that the small number of images in each SOP class (5.26 images on average) causes inaccurate parameter estimation for von Mises-Fisher (vMF) distributions. To verify this hypothesis, we examine the estimation error in the $\kappa$ parameters from different amount of data. We generate synthetic data points from a ground-truth vMF distribution and report the estimation error in Figure~\ref{fig:vMF_mse}. When there are only 5 samples per class, the MSE is approximately 155.6. When we have more than 34 samples, the error drops below 20. High errors under small data lead to inaccurate estimation of $P_{\text{clean}}(i)$ and explains the degraded performance of vMF-Sim on SOP. 
Similarly, the proxy-based method may not learn accurate proxies due to limited training data per class. 
The results suggest that AvgSim is the better option than vMF-Sim or ProxySim when the number of samples per class is small.

\begin{figure}[t!]
    \centering
    \includegraphics[width=0.7\columnwidth]{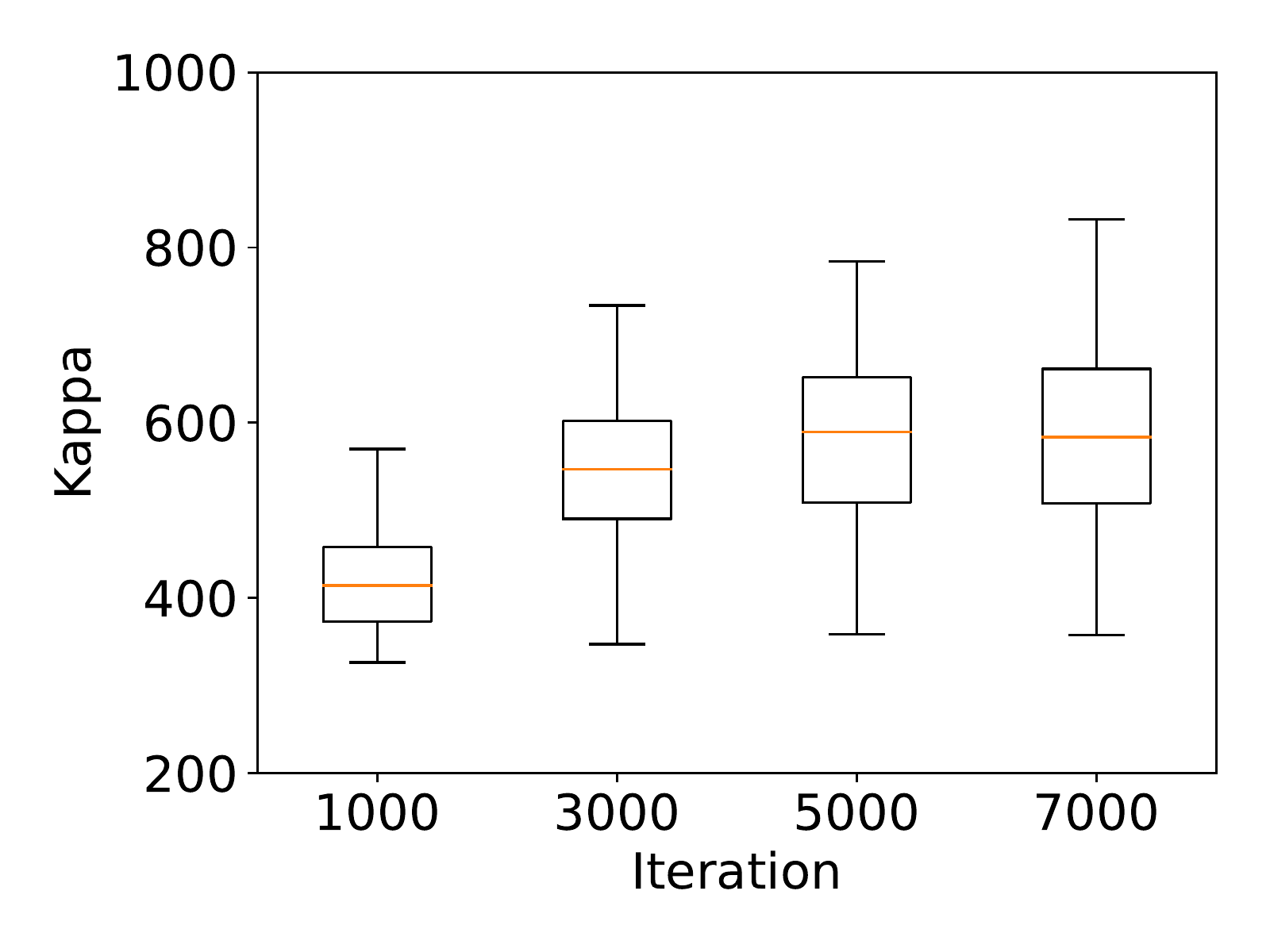}
    \caption{Box plot of $\kappa$ as training progresses, calculated on CUB with 50\% Symmetric Noise and SoftTriple loss. Increasing $\kappa$ indicates same-class representations become more concentrated as training progresses. }
    \label{fig:kappa}
\end{figure}

\vspace{0.1in} \noindent \textbf{Dynamics of $\kappa$ in vMF-Sim}. To understand the dynamic behaviors of vMF-Sim over the course of training, we show the distributions of $\kappa$ estimates for all classes at 1k, 3k, 5k and 7k training iterations in Figure~\ref{fig:kappa}. $\kappa$ is relatively small at the 1,000th iteration and gradually increases in subsequent training iterations. This suggests that the data representations within a class become increasingly concentrated on their centers as training progresses, as the network learns to pull data points within the same class together. This result also demonstrates that $\kappa$ values are not equal across classes and training iterations, which was the assumption adopted by existing works \cite{hasnat2017mises,zhe2019directional}.

\subsection{Ablation Studies}
\begin{figure}[]
      \centering
     
      \includegraphics[width=0.8\columnwidth]{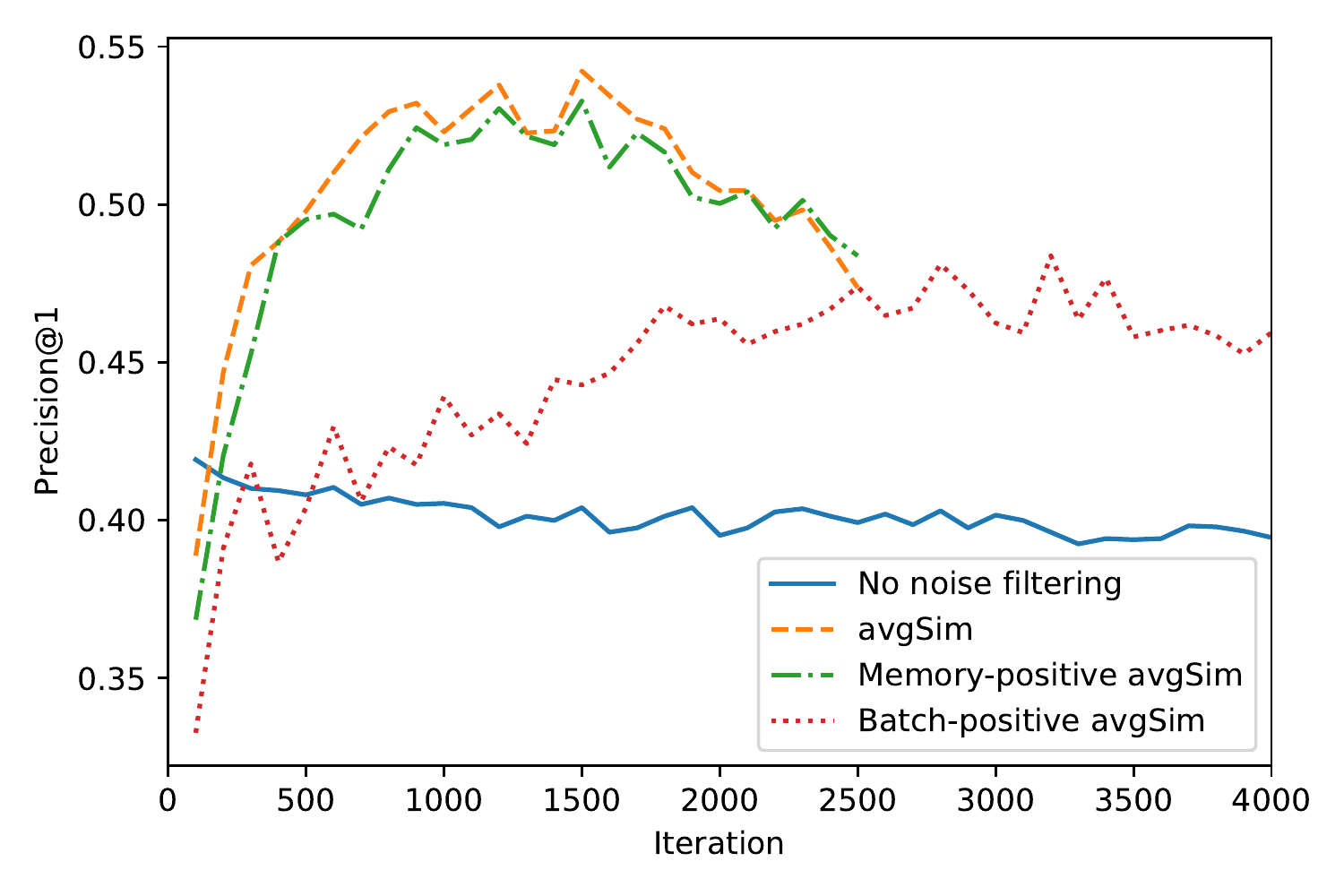}
      	\caption{Precision@1 (\%) vs number of iterations on CUB with 50\% Symmetric Noise. The performance on validation set is reported. The training of AvgSim and Memory-positive AvgSim is early stopped.}\label{fig:compare_avgSim}
\end{figure}

\noindent \textbf{Design of $P_{\text{clean}}(i)$.} We compare two alternative designs of the clean-data identification function $P_{\text{clean}}(i)$ for average similarity method (AvgSim). The first design, namely Batch-positive AvgSim, uses only the average similarity of the positive pairs in the current minibatch. The negative pairs are not utilized.
\begin{equation}\label{equ:ablation_design_noMemory}
    P_{\text{clean}}(i) \propto \frac{1}{K-1}\sum_{(x_j,y_j) \in \mathcal{B}, y_j=y_i } S(f(x_i),f(x_j)),
\end{equation}
The second baseline, Memory-positive, uses the average similarity between $x_i$ and other same-class samples retrieved from the memory bank, but still does not consider the negative samples.
\begin{equation}\label{equ:ablation_design_withMemory}
    P_{\text{clean}}(i) \propto \frac{1}{M_{y_i}}\sum_{(v_j,y_j) \in \mathcal{M}, \,y_j=y_i}S(f(x_i),v_j).
\end{equation}
Figure~\ref{fig:compare_avgSim} shows that Precision@1 changes as the training iteration increases. We use Memory Contrastive Loss (MCL) \cite{wang2020cross} as the loss function. Due to high levels of noise, training with no noise filtering identifies only 25\% correct pairs, which makes performance deteriorate right from the beginning phase of training (Figure~\ref{fig:compare_avgSim}). All filtering strategies improve performance relative to the scenario of no noise filtering. The Memory-positive strategy works better than Batch-positive, showing the importance of using the entire memory bank to identify noise. AvgSim achieves the best P@1 scores. 

\begin{figure}[]
      \centering

      \includegraphics[width=0.6\columnwidth]{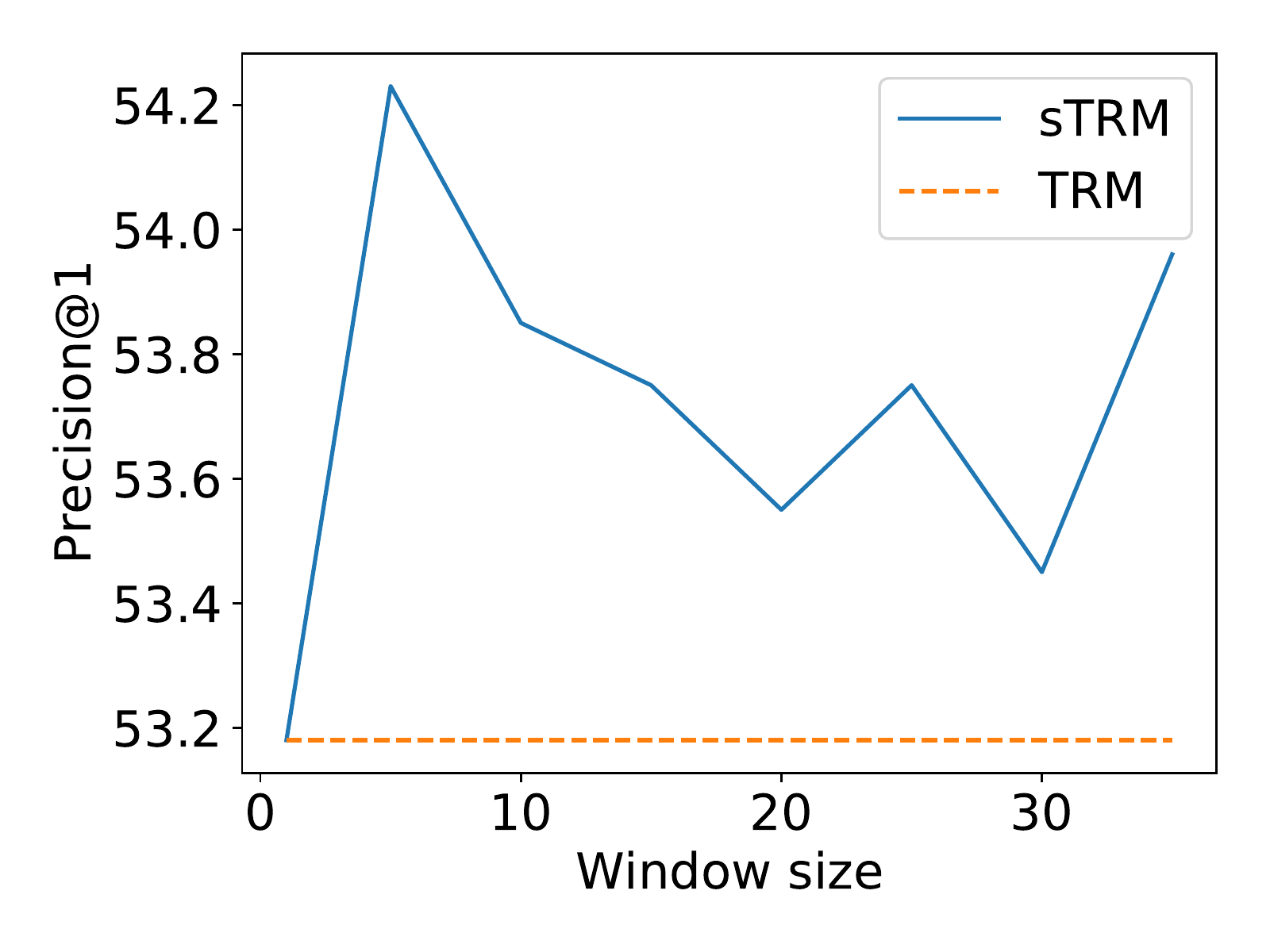}
      \caption{Precision@1 (\%) vs. window size $\tau$ on the CUB validation set. The model is trained with SoftTriple loss and AvgSim under 50\% symmetric noise.}\label{fig:window_size}
\end{figure}

\vspace{0.1in}
\noindent \textbf{Clean data threshold in AvgSim.}
We compare the two different methods, sTRM and TRM, for setting the threshold for $P_{\text{clean}}(i)$ in AvgSim. 
Figure~\ref{fig:window_size} illustrates the model performance for different choices of the sliding window size in sTRM. Note that TRM is a special case of sTRM, where the window size $\tau$ is set to $1$. Across all different choices of $\tau$, sTRM consistently outperforms TRM. 

\section{Conclusions and Future Work}
In this paper, we propose a novel, efficient, simple and effective approach, Probabilistic Ranking-based Instance Selection with Memory (PRISM), to enhance the performance of deep metric learning in the presence of training label noise. It calculates the probability of a data point being clean, $P_{\text{clean}}(i)$, and uses it to filter out noisy data. 
We introduce three methods to calculate $P_{\text{clean}}(i)$, namely, average similarity method (AvgSim), proxy similarity method (ProxySim) and von Mises-Fisher Distribution Similarity (vMF-Sim). AvgSim uses the average similarity between potentially noisy data and clean data. ProxySim replaces the centers maintained by AvgSim with the proxy trained by proxy-based method. 
vMF-Sim considers not only the similarity between data and classes, but also the ``tightness'' of the data distribution of each class.
Through extensive experiments with both synthetic and real-world datasets, we demonstrate that PRISM significantly outperforms existing approaches. Our analysis of the proposed similarity evaluation methods reveal their characteristics in terms of prediction accuracy and computational time so that practitioners can make informed decisions when selecting the most suitable technique.
PRISM holds the promise to enable DML approach to be applied in diverse machine learning applications with significant levels of data noise. 

The PRISM family of approaches are only the first step into enabling DML approach to be robust in the presence of label noise. Currently, they can only identify noisy data and exclude them from the training process. An interesting future work is to enable label correction after identifying the noisy labels in DML, such that the noisy data can also be leveraged for model training. 

\section*{Acknowledgment}
We gratefully acknowledge the support by the National Research Foundation, Singapore through the AI Singapore Programme (AISG2-RP-2020-019), NRF Investigatorship (NRF-NRFI05-2019-0002), and NRF Fellowship (NRF-NRFF13-2021-0006); Alibaba Group through Alibaba Innovative Research and Alibaba-NTU Singapore Joint Research Institute (Alibaba-NTU-AIR2019B1); the Nanyang Assistant/Associate Professorships; NTU-SDU-CFAIR (NSC-2019-011); NSFC No.91846205; the Innovation Method Fund of China No.2018IM020200; the RIE 2020 Advanced Manufacturing and Engineering Programmatic Fund (No. A20G8b0102), Singapore. 

\bibliographystyle{IEEEtran}
\bibliography{bib}
 
\vspace{11pt}
\begin{IEEEbiography}[{\includegraphics[width=1in,height=1.25in,clip,keepaspectratio]{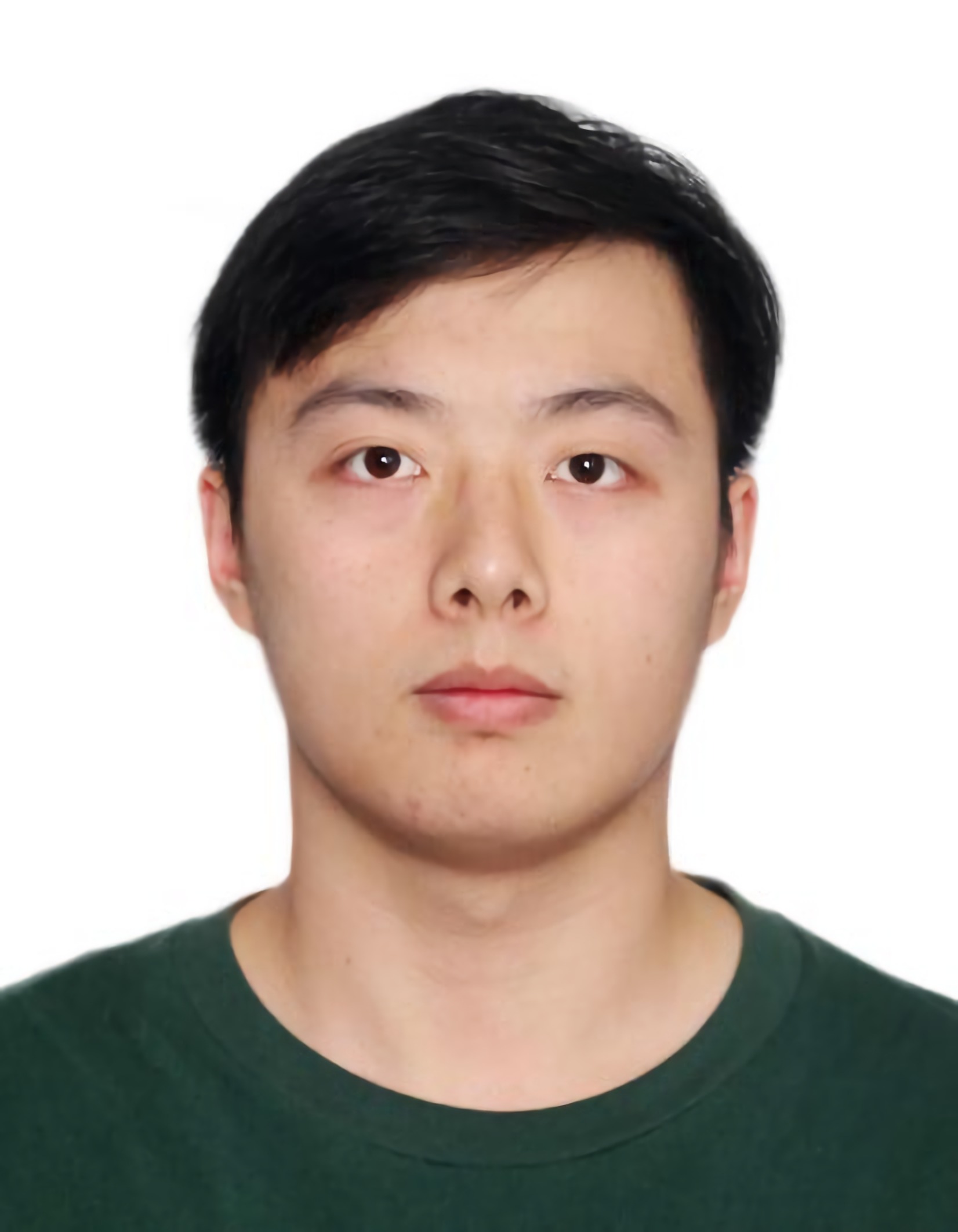}}]{Chang Liu}
received the BS degree in School of Computer Science and Technology, Shandong University, China, in 2018. He is currently a PhD student in the School of Computer Science and Engineering (SCSE), Nanyang Technological University (NTU), Singapore. His research interests include deep metric learning and AI-empowered video generation.
\end{IEEEbiography}
\begin{IEEEbiography}[{\includegraphics[width=1in,clip,keepaspectratio]{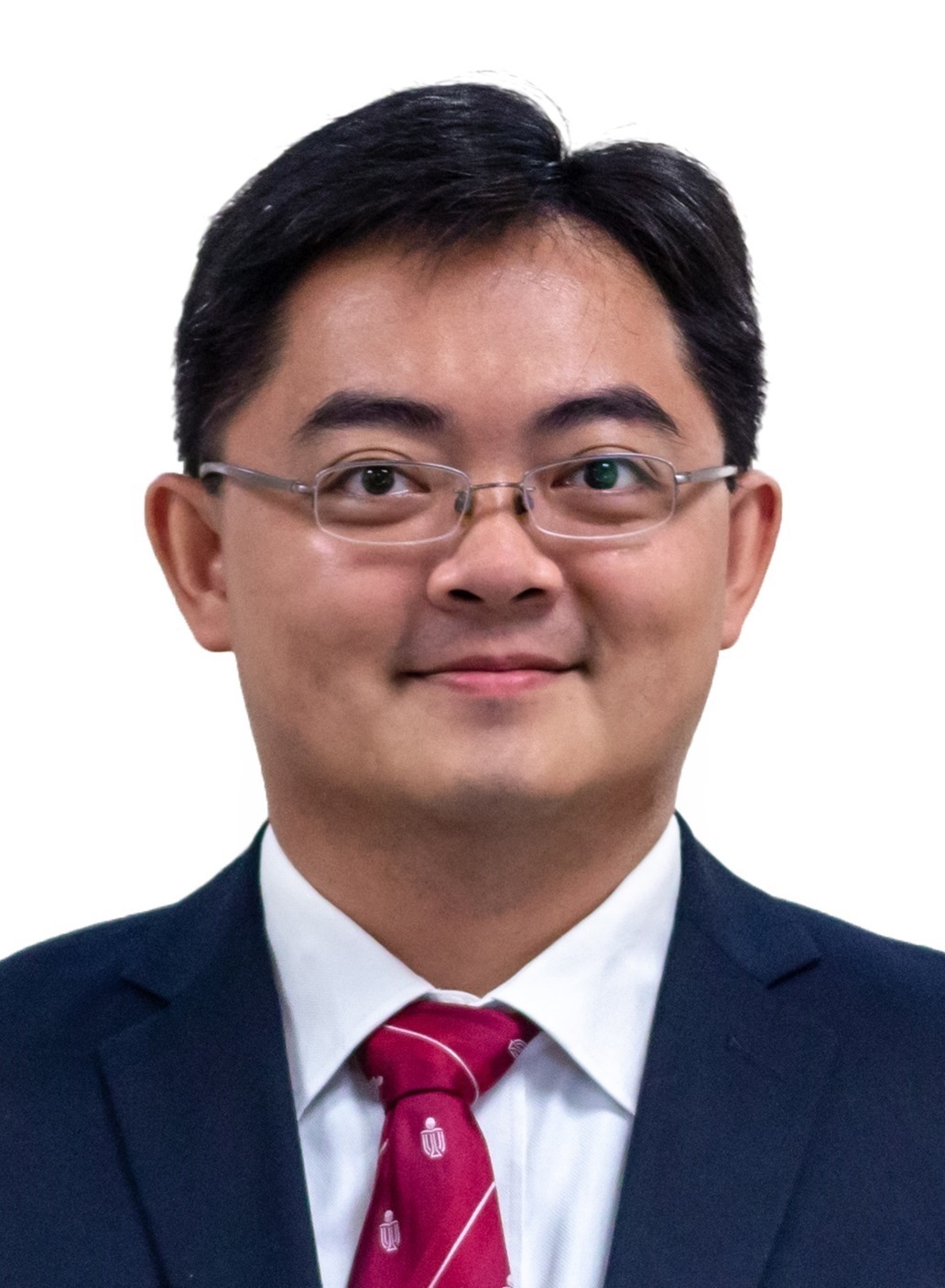}}]{Han Yu}
	is a Nanyang Assistant Professor in the School of Computer Science and Engineering (SCSE), Nanyang Technological University (NTU), Singapore. He has been a Visiting Scholar at the Department of Computer Science and Engineering, Hong Kong University of Science and Technology. He received his PhD and B.Eng (Hons) from SCSE, NTU in 2014 and 2007, respectively. His research focuses on trustworthy AI and federated learning. His works have been recognized with multiple awards.
\end{IEEEbiography}
\begin{IEEEbiography}[{\includegraphics[width=1in,clip,keepaspectratio]{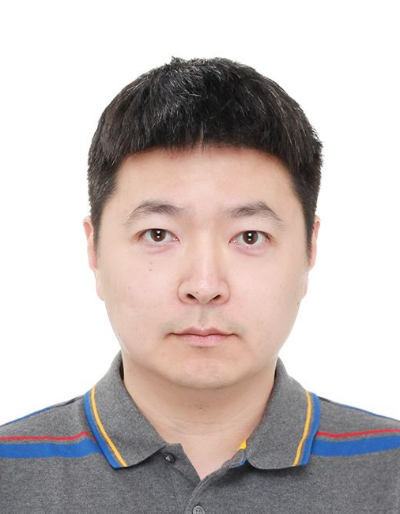}}]{Boyang Li}
	is a Nanyang Associate Professor in the School of Computer Science and Engineering, Nanyang Technological University, Singapore. Previously, he was a Senior Research Scientist at Baidu Research USA and a Research Scientist at Disney Research. He received his Ph.D. in Computer Science from Georgia Institute of Technology. His current research interests include Machine Learning, Multimodal Learning, and Computational Narrative Intelligence. 
\end{IEEEbiography}
\begin{IEEEbiography}[{\includegraphics[width=1in,clip,keepaspectratio]{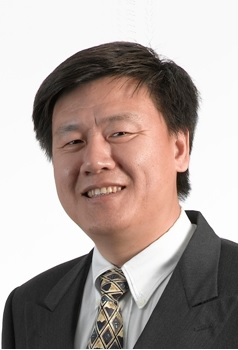}}]{Zhiqi Shen}
	is currently a Senior Lecturer in the School of Computer Science and Engineering, Nanyang Technological University (NTU), Singapore. He obtained a BSc degree in Computer Science from Peking University, a Master’s degree from Beijing University of Technology and a PhD degree from NTU, respectively. 
    His research interests include goal oriented intelligent agents, multi agent systems, agent oriented software engineering and interdisciplinary research in artificial intelligence (AI), machine learning for computer vision, game design, digital storytelling, e-learning, e-health, crowdsourcing and active ageing. 
    He has published over 200 research papers and has led over 20 research projects that are commercialised and translated to real world systems. 
\end{IEEEbiography}
\begin{IEEEbiography}[{\includegraphics[width=1in,clip,keepaspectratio]{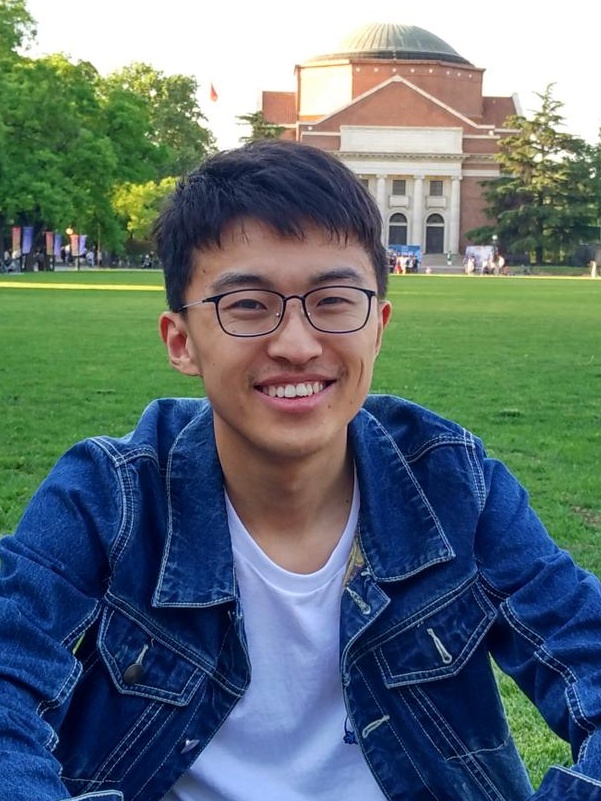}}]{Zhanning Gao}
	received the B.S. and Ph.D. degree in Control Science and Engineering from Xi’an Jiaotong University in 2012 and 2018. He was a research intern in Visual Computing Group in Microsoft Research Asia from 2015 to 2017. He is currently a senior algorithm expert of Alibaba DAMO Academy. His research interests include compact image/video representation, complex event video analysis and high-resolution video synthesis.
\end{IEEEbiography}
\begin{IEEEbiography}[{\includegraphics[width=1in,clip,keepaspectratio]{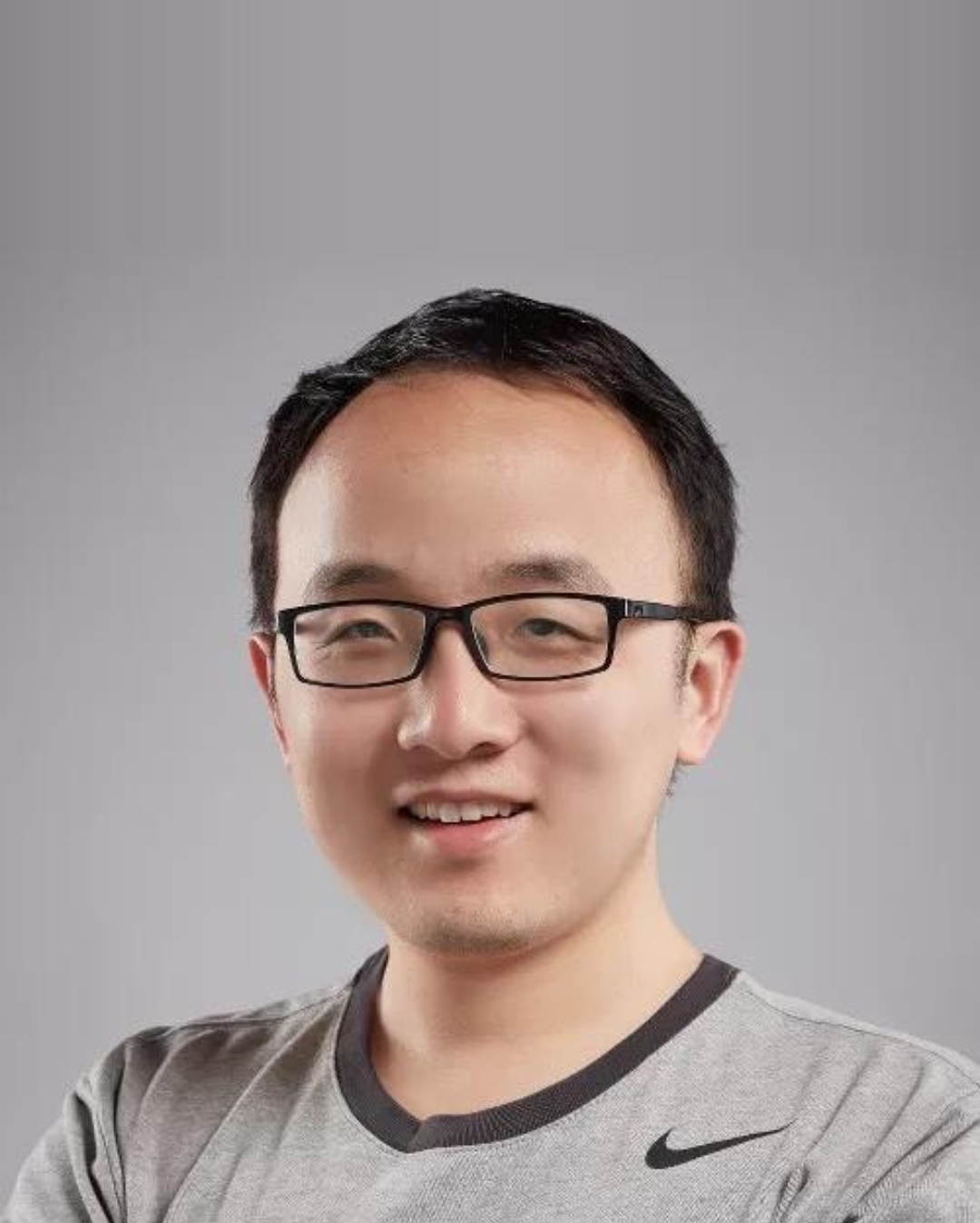}}]{Peiran Ren}
	received his BSc and PhD degree from Tsinghua University, China, in 2008 and 2014 respectively. He is now a senior algorithm engineer at Alibaba Damo Acadamy. His research interests include image and video enhancement and processing, computer aided design, real-time rendering, and appearance acquisition.
\end{IEEEbiography}
\begin{IEEEbiography}[{\includegraphics[width=1in,clip,keepaspectratio]{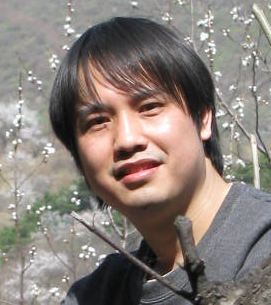}}]{Xuansong Xie}
	received PhD degree from JiLin University, China, in 2006. He is now a senior algorithm expert of Alibaba DAMO Academy. His research interests include image retrival, computer aided design, visual understanding and production. Now he's in charge of Alibaba visual intelligence open platform.
\end{IEEEbiography}
\begin{IEEEbiography}[{\includegraphics[width=1in,clip,keepaspectratio]{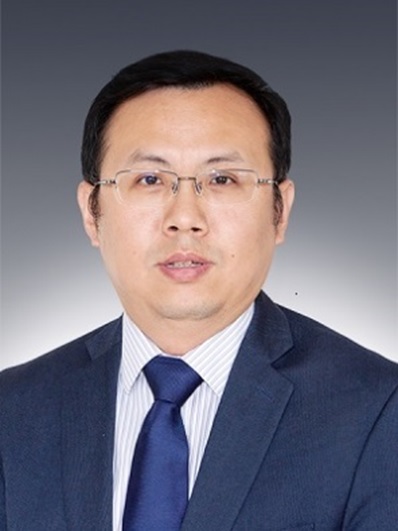}}]{Lizhen Cui}
	is a Professor and the Vice Chair of the School of Software, Shandong University. Between 2013 and 2014, he was a Visiting Scholar at the Georgia Institute of Technology. His main research interests include data science and engineering, intelligent data analysis, service computing and collaborative computing.
\end{IEEEbiography}

\vfill

\end{document}